\pdfoutput=1
\documentclass[11pt, a4paper, logo, copyright, nonumbering]{hku}
\usepackage{natbib}

\usepackage[utf8]{inputenc} 
\usepackage[T1]{fontenc}    
\usepackage{booktabs}       
\usepackage{amsfonts}       
\usepackage{nicefrac}       
\usepackage{microtype}      
\usepackage{xcolor}         

\usepackage{graphicx}
\usepackage{multirow}
\usepackage{enumitem}
\usepackage{subcaption}
\usepackage{xspace}
\usepackage{makecell}
\usepackage{tcolorbox}

\usepackage{fontawesome5}  
\usepackage{array}
\definecolor{mydarkblue}{rgb}{0,0.08,0.45}
\definecolor{linkcolor}{rgb}{0.956,0.298,0.235}
\definecolor{citecolor}{HTML}{1976D2}
\hypersetup{
    colorlinks=true,
    linkcolor=citecolor,
    filecolor=magenta,
    urlcolor=cyan,
    citecolor=citecolor,
}

\usepackage{rotating}
\usepackage{enumitem}
\usepackage[capitalize,noabbrev]{cleveref}
\crefname{section}{Sec.}{Secs.}
\Crefname{section}{Section}{Sections}
\Crefname{table}{Table}{Tables}
\crefname{table}{Tab.}{Tabs.}

\newcommand{\ours}{\mbox{Goku}\xspace}

\newcommand{\eg}{\textit{e}.\textit{g}.}

\definecolor{deemph}{gray}{0.6}

\newcommand{\tablestyle}[2]{\setlength{\tabcolsep}{#1}\renewcommand{\arraystretch}{#2}\centering}

\reportnumber{001}

\renewcommand{\today}{}

\title{\centering Goku: Flow Based Video Generative Foundation Models}

\author{%
  \textcolor{red}{TBD} \\
}

\author[*]{
\small
Shoufa Chen$^{1*}$ \quad Chongjian Ge$^{1*}$ \quad Yuqi Zhang$^2$ \quad Yida Zhang$^2$ \quad Fengda Zhu$^2$ \quad Hao Yang$^2$ \quad Hongxiang Hao$^2$ \quad Hui Wu$^2$ \quad Zhichao Lai$^2$ \quad Yifei Hu$^2$ \quad Ting-Che Lin$^2$ \quad Shilong Zhang$^1$ \quad Fu Li$^2$ \quad Chuan Li$^2$ \quad Xing Wang$^2$ \quad Yanghua Peng$^2$ \quad Peize Sun$^{1}$ \quad Ping Luo$^{1}$ \quad Yi Jiang$^2$ \quad Zehuan Yuan$^2$ \quad Bingyue Peng$^2$ \quad Xiaobing Liu$^2$ \\

\small
$^1$The University of Hong Kong \quad $^2$Bytedance Inc \\
\small
$^*$ Equal Contribution \\ 
\quad \\
\small
\url{https://saiyan-world.github.io/goku/}
}

\begin{abstract}
This paper introduces \textbf{\emph{\ours}}, a state-of-the-art family of joint image-and-video generation models leveraging rectified flow Transformers to achieve industry-leading performance. We detail the foundational elements enabling high-quality visual generation, including the data curation pipeline, model architecture design, flow formulation, and advanced infrastructure for efficient and robust large-scale training.
The \ours models demonstrate superior performance in both qualitative and quantitative evaluations, setting new benchmarks across major tasks. Specifically, \ours achieves 0.76 on GenEval and 83.65 on DPG-Bench for text-to-image generation, and 84.85 on VBench for text-to-video tasks. We believe that this work provides valuable insights and practical advancements for the research community in developing joint image-and-video generation models.
\end{abstract}

\begin{document}

\maketitle

\section{Introduction}

Video generation has garnered significant attention owing to its transformative potential across a wide range of applications, such media content creation~\citep{polyak2024movie}, advertising~\citep{zhang2024virbo,bacher2021advert}, video games~\citep{yang2024playable,valevski2024diffusion, oasis2024}, and world model simulators~\citep{ha2018world, videoworldsimulators2024, agarwal2025cosmos}. Benefiting from advanced generative algorithms~\citep{goodfellow2014generative, ho2020denoising, liu2023flow, lipman2023flow}, scalable model architectures~\citep{vaswani2017attention, peebles2023scalable}, vast amounts of internet-sourced data~\citep{chen2024panda, nan2024openvid, ju2024miradata}, and ongoing expansion of computing capabilities~\citep{nvidia2022h100, nvidia2023dgxgh200, nvidia2024h200nvl}, remarkable advancements have been achieved in the field of video generation~\citep{ho2022video, ho2022imagen, singer2023makeavideo, blattmann2023align, videoworldsimulators2024, kuaishou2024klingai, yang2024cogvideox, jin2024pyramidal, polyak2024movie, kong2024hunyuanvideo, ji2024prompt}.

In this work, we present \textbf{\ours}, a family of rectified flow~\citep{lipman2023flow, liu2023flow} transformer models designed for joint image and video generation, establishing a pathway toward industry-grade performance. This report centers on four key components: data curation, model architecture design, flow formulation, and training infrastructure optimization—each rigorously refined to meet the demands of high-quality, large-scale video generation.

\begin{figure}[ht]
    \centering
    \begin{subfigure}[b]{0.82\linewidth}
        \centering
        \includegraphics[width=\linewidth]{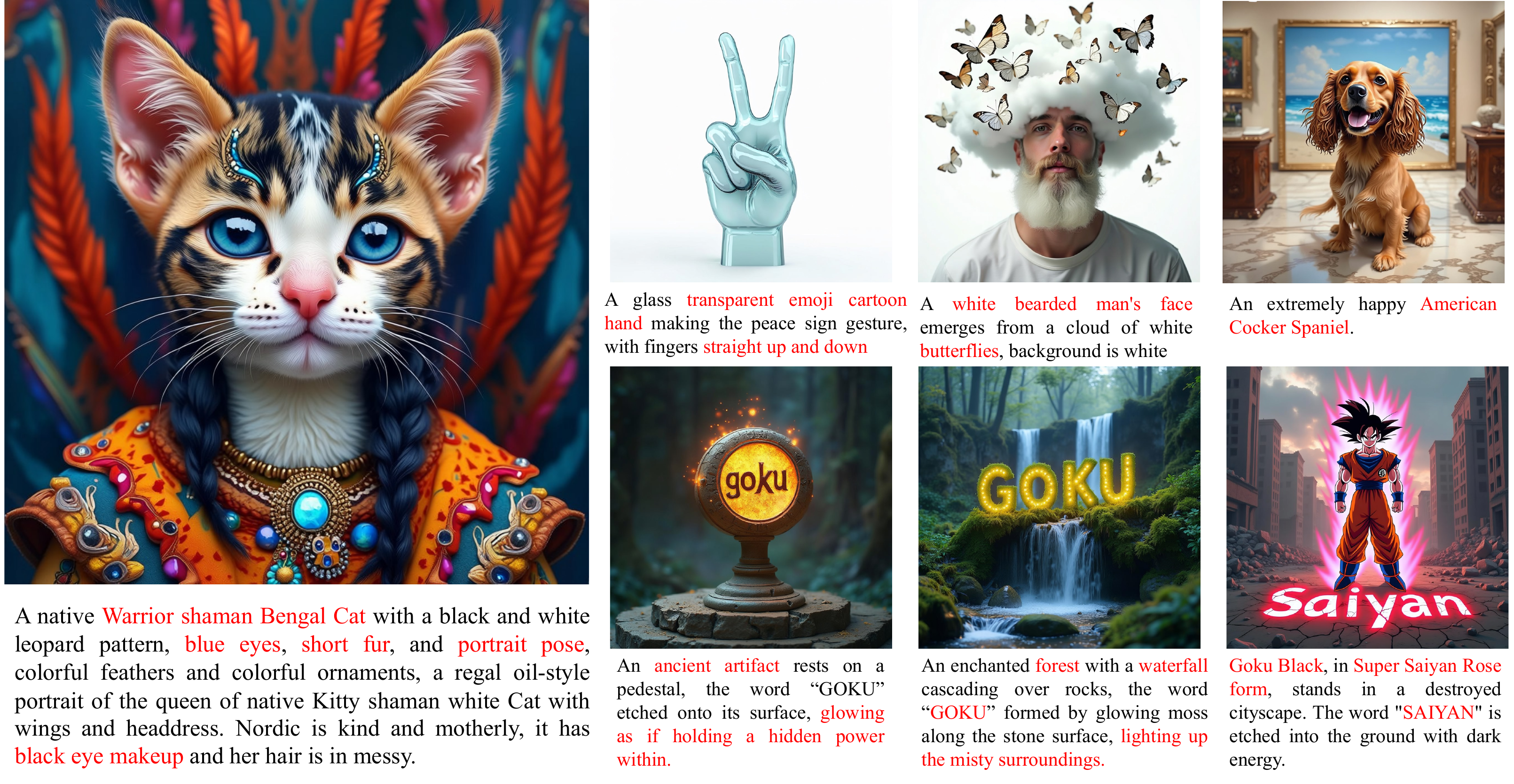}
        \caption{Text-to-Image Samples}\label{fig:main-demo-t2i}
    \end{subfigure}
    \vfill
    \begin{subfigure}[b]{0.82\linewidth}
        \centering
        \includegraphics[width=\linewidth]{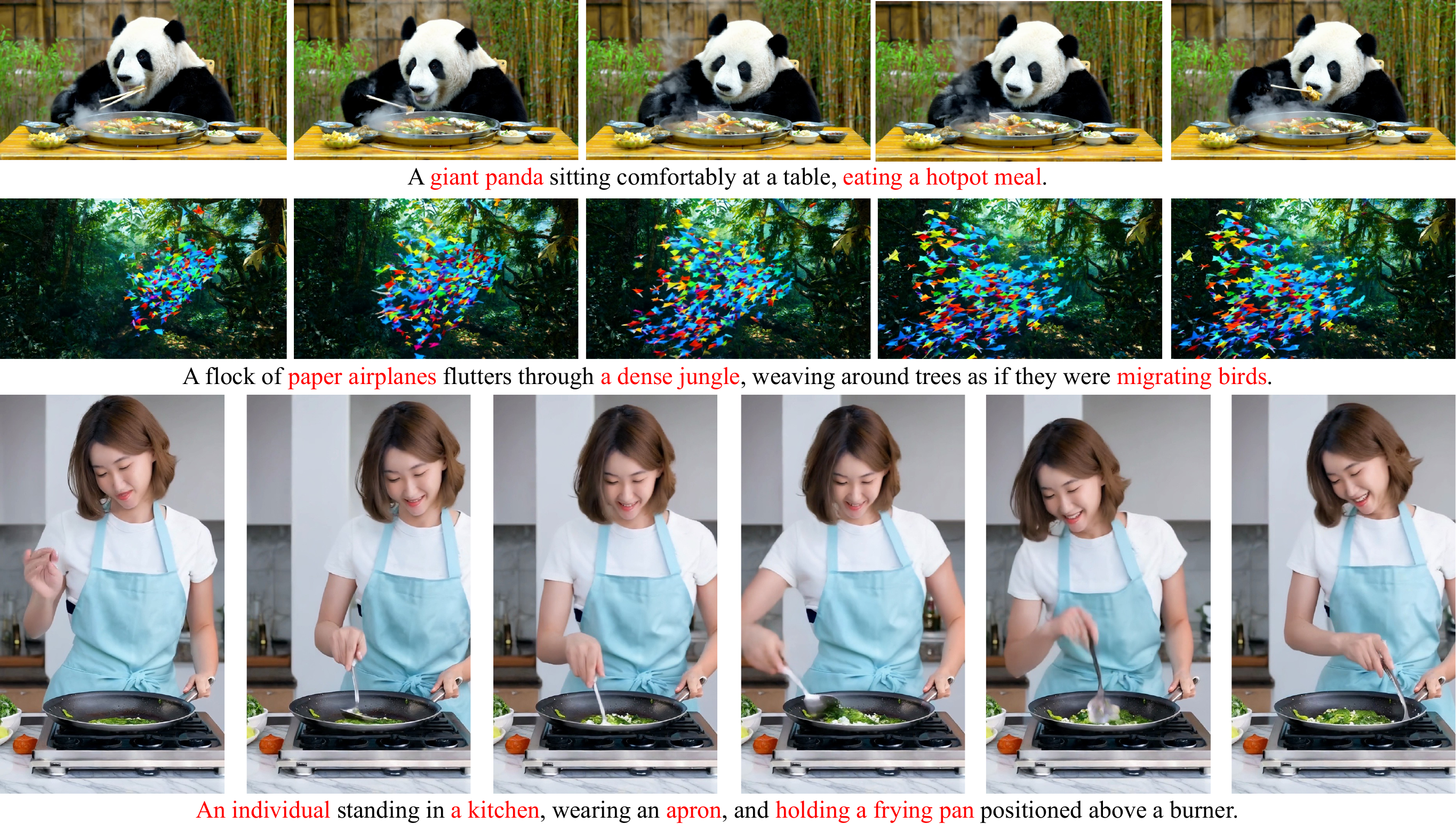}
        \caption{Text-to-Video Samples}\label{fig:main-demo-t2v}
    \end{subfigure}
\caption{\textbf{Generated samples from \ours.} Key components are highlighted in \textcolor{red}{\textbf{RED}}.}\label{fig:main-demo}
\end{figure}

First, we present a comprehensive data processing pipeline designed to construct large-scale, high-quality image and video-text datasets. The pipeline integrates multiple advanced techniques, including video and image filtering based on aesthetic scores, OCR-driven content analysis, and subjective evaluations, to ensure exceptional visual and contextual quality. Furthermore, we employ multimodal large language models~(MLLMs)~\citep{yuan2025tarsier2} to generate dense and contextually aligned captions, which are subsequently refined using an additional large language model~(LLM)~\citep{yang2024qwen2} to enhance their accuracy, fluency, and descriptive richness. As a result, we have curated a robust training dataset comprising approximately 36M video-text pairs and 160M image-text pairs, which are proven sufficient for training industry-level generative models.

Secondly, we take a pioneering step by applying rectified flow formulation~\citep{lipman2023flow} for joint image and video generation, implemented through the \ours model family, which comprises Transformer architectures with 2B and 8B parameters. At its core, the \ours framework employs a 3D joint image-video variational autoencoder (VAE) to compress image and video inputs into a shared latent space, facilitating unified representation. This shared latent space is coupled with a full-attention~\citep{vaswani2017attention} mechanism, enabling seamless joint training of image and video. This architecture delivers high-quality, coherent outputs across both images and videos, establishing a unified framework for visual generation tasks.

Furthermore, to support the training of \ours at scale, we have developed a robust infrastructure tailored for large-scale model training. Our approach incorporates advanced parallelism strategies~\citep{jacobs2023deepspeed, pytorch_fsdp} to manage memory efficiently during long-context training. Additionally, we employ ByteCheckpoint~\citep{wan2024bytecheckpoint} for high-performance checkpointing and integrate fault-tolerant mechanisms from MegaScale~\citep{jiang2024megascale} to ensure stability and scalability across large GPU clusters. These optimizations enable \ours to handle the computational and data challenges of generative modeling with exceptional efficiency and reliability.

We evaluate \ours on both text-to-image and text-to-video benchmarks to highlight its competitive advantages. For text-to-image generation, \ours-T2I demonstrates strong performance across multiple benchmarks, including T2I-CompBench~\citep{huang2023t2i-compbench}, GenEval~\citep{ghosh2024geneval}, and DPG-Bench~\citep{hu2024ella_dbgbench}, excelling in both visual quality and text-image alignment. In text-to-video benchmarks, \ours-T2V achieves state-of-the-art performance on the UCF-101~\citep{ucf101} zero-shot generation task. Additionally, \ours-T2V attains an impressive score of \textbf{84.85} on VBench~\citep{huang2024vbench}, securing the top position on the leaderboard (as of 2025-01-25) and surpassing several leading commercial text-to-video models. Qualitative results, illustrated in \Cref{fig:main-demo}, further demonstrate the superior quality of the generated media samples. These findings underscore \ours's effectiveness in multi-modal generation and its potential as a high-performing solution for both research and commercial applications.
\section{\ours: Generative Flow Models for Visual Creation}

In this section, we present three core components of \ours, the image-video joint VAE~\citep{yang2024cogvideox}, the \ours Transformer architecture, and the rectified flow formulation. These components are designed to work synergistically, forming a cohesive and scalable framework for joint image and video generation. During training, each raw video input $x\in \mathbb{R}^{T\times H \times W \times3}$ (with images treated as a special case where $T=1$ ) is encoded from the pixel space to a latent space using a 3D image-video joint VAE~(\Cref{sec:vae}). The encoded latents are then organized into mini-batches containing both video and image representations, facilitating the learning of a unified cross-modal representation. Subsequently, the rectified flow formulation~(\Cref{sec:method-flow}) is applied to these latents, leveraging a series of Transformer blocks~(\Cref{sec:trans-block}) to model complex temporal and spatial dependencies effectively.

\subsection{Image-Video Joint VAE}\label{sec:vae}
Earlier research \citep{he2022latent,rombach2022high,esser2021taming} demonstrates that diffusion and flow-based models can significantly improve efficiency and performance by modeling in latent space through a Variational Auto-Encoder (VAE)~\citep{esser2021taming, kingma2013auto}. Inspired by Sora~\citep{videoworldsimulators2024}, the open-source community has introduced 3D-VAE to explore spatio-temporal compression within latent spaces for video generation tasks~\citep{pku_yuan_lab_and_tuzhan_ai_etc_2024_10948109, opensora, yang2024cogvideox}. To extend the advantages of latent space modeling across multiple media formats, including images and videos, we adopt a jointly trained Image-Video VAE~\citep{yang2024cogvideox} that handles both image and video data within a unified framework. Specifically, for videos, we apply a compression stride of $8 \times 8 \times 4$ across height, width, and temporal dimensions, respectively, while for images, the compression stride is set to $8 \times 8$ in spatial dimensions.

\subsection{Transformer Architectures}\label{sec:trans-block}

The design of the \ours Transformer block builds upon GenTron~\citep{chen2024gentron}, an extension of the class-conditioned diffusion transformer~\citep{peebles2023scalable} for text-to-image/video tasks. It includes a self-attention module for capturing inter-token correlations, a cross-attention layer to integrate textual conditional embeddings (extracted via the Flan-T5 language model~\citep{chung2024scaling}), a feed-forward network~(FFN) for feature projection, and a layer-wise adaLN-Zero block that incorporates timestep information to guide feature transformations. Additionally, we introduce several recent design enhancements to improve model performance and training stability, as detailed below.

\begin{table}[t]
    \centering
    \tablestyle{8pt}{1.15}
    \begin{tabular}{ccccc}
    \toprule
    Model & Layer & Model Dimension & FFN Dimension & Attention Heads  \\
    \midrule
    \ours-1B & 28 & 1152 & 4608  & 16 \\
    \ours-2B & 28 & 1792 & 7168  & 28 \\
    \ours-8B & 40 & 3072 & 12288 & 48 \\
    \bottomrule
    \end{tabular}
    \caption{\textbf{Architecture configurations for \ours Models.} \ours-1B model is only used for pilot experiments in \Cref{sec:method-flow}}\label{tab:model-config}
\end{table}

\paragraph{Plain Full Attention.} In Transformer-based video generative models, previous approaches~\citep{chen2024gentron, wu2023tune, singer2023makeavideo, blattmann2023align} typically combine temporal attention with spatial attention to extend text-to-image generation to video. While this method reduces computational cost, it is sub-optimal for modeling complex temporal motions, as highlighted in prior work~\citep{yang2024cogvideox, polyak2024movie}. In \ours, we adopt full attention to model multi-modal tokens (image and video) within a unified network. Given the large number of video tokens remaining after VAE processing—particularly for high-frame-rate, long-duration videos—we leverage \texttt{FlashAttention}~\citep{shah2024flashattention, dao2023flashattention2} and sequence parallelism~\citep{li2021sequence} to optimize both GPU memory usage and computational efficiency.

\paragraph{Patch n’ Pack.} To enable joint training on images and videos of varying aspect ratios and lengths, we follow the approach from NaViT~\citep{dehghani2024patch}, packing both modalities into a single minibatch along the sequence dimension. This method allows flexible mixing of training instances with different sequence lengths into a single batch, eliminating the need for data buckets~\citep{podell2023sdxl}.

\paragraph{3D RoPE Position Embedding.} Rotary Position Embedding (RoPE)~\citep{su2024roformer} has demonstrated effectiveness in LLMs by enabling greater sequence length flexibility and reducing inter-token dependencies as relative distances increase. During joint training, we apply 3D RoPE embeddings to image and video tokens. In our joint training framework, we extend 3D RoPE embeddings to image and video tokens, leveraging their extrapolation capability to accommodate varying resolutions. This adaptability makes RoPE particularly suited for handling diverse resolutions and video lengths. Furthermore, our empirical analysis revealed that RoPE converges faster than sinusoidal positional embeddings during transitions across different training stages

\paragraph{Q-K Normalization.} Training large-scale Transformers can occasionally result in loss spikes, which may lead to model corruption, manifesting as severe artifacts or even pure noise in generated images or videos. To mitigate this issue, we incorporate query-key normalization~\citep{dehghani2023scaling} to stabilize the training process. Specifically, we apply RMSNorm~\citep{zhang2019root} to each query-key feature prior to attention computation, ensuring smoother and more reliable training dynamics.

The overall Transformer model is constructed by stacking a sequence of blocks as described above. To address varying computational demands and performance requirements, we design three model variants, summarized in \Cref{tab:model-config}. The \ours-1B model serves as a lightweight option for pilot experiments. The \ours-2B variant consists of 28 layers, each with a model dimension of 1792 and 28 attention heads, providing a balance between computational efficiency and expressive capacity. In contrast, the larger \ours-8B variant features 40 layers, a model dimension of 3072, and 48 attention heads, delivering superior modeling capacity aimed at achieving high generation quality.

\subsection{Flow-based Training}\label{sec:method-flow}
Our flow-based formulation is rooted in the rectified flow~(RF) algorithm~\citep{albergo2023building, lipman2023flow, liu2023flow}, where a sample is progressively transformed from a prior distribution, such as a standard normal distribution, to the target data distribution. This transformation is achieved by defining the forward process as a series of linear interpolations between the prior and target distributions. Specifically, given a real data sample $\mathbf{x}_1$ from the target distribution and a noise sample $\mathbf{x}_0 \sim \mathcal{N}(0, 1)$ from the prior distribution, a training example is constructed through linear interpolation:

\begin{equation}\label{eq:flow-formulation}
\mathbf{x}_t = t \cdot \mathbf{x}_1 + (1 - t) \cdot \mathbf{x}_0,
\end{equation}

\noindent where $t \in [0, 1]$ represents the interpolation coefficient. The model is trained to predict the velocity, defined as the time derivative of $\mathbf{x}_t$, $\mathbf{v}_t = \frac{d\mathbf{x}_t}{dt}$, which guides the transformation of intermediate samples $\mathbf{x}_t$ towards the real data $\mathbf{x}_1$ during inference. By establishing a direct, linear interpolation between data and noise, RF simplifies the modeling process, providing improved theoretical properties, conceptual clarity, and faster convergence across data distributions.

\ours takes a pioneering step by adopting a flow-based formulation for joint image-and-video generation.
We conduct a pilot experiment to validate the rapid convergence of flow-based training by performing class-conditional generation with \ours-1B a model specifically designed for these proof-of-concept experiments, on ImageNet-1K ($256 \times 256$)~\citep{deng2009imagenet}. The model is configured with 28 layers, an attention dimension of 1152, and 16 attention heads. To evaluate performance, we compare key metrics, such as FID-50K and Inception Score~(IS), for models trained using the denoising diffusion probabilistic model (DDPM)~\citep{ho2020denoising} and rectified flow.  As shown in \Cref{tab:toy-fid}, RF demonstrates  faster convergence than DDPM. For instance, \ours-1B~(RF) achieves a lower FID-50K after 400k training steps compared to \ours-1B~(DDPM), which requires 1000k steps to reach a similar level of performance.

\subsection{Training Details}\label{sec:training}
\paragraph{Multi-stage Training.} Directly optimizing joint image-and-video training poses significant challenges, as the network must simultaneously learn spatial semantics critical for images and temporal motion dynamics essential for videos. To tackle this complexity, we introduce a decomposed, multi-stage training strategy that progressively enhances the model’s capabilities, ensuring effective and robust learning across both modalities.

\begin{table}[t]
    \centering
    \tablestyle{9pt}{1.12}
    \begin{tabular}{cccccccc}
    \toprule
    Loss & Steps & FID $\downarrow$ & sFID $\downarrow$ & IS $\uparrow$ & Precision $\uparrow$ & Recall $\uparrow$ \\
    \midrule
    DDPM & 200k & 3.0795 & 4.3498 & 226.4783 & 0.8387 & 0.5317  \\
    DDPM & 400k & 2.5231 & 4.3821 & 265.0612 & 0.8399 & 0.5591 \\
    DDPM & 1000k & 2.2568 & 4.4887 & 286.5601 & 0.8319 & 0.5849  \\
    \midrule
    Rectified Flow & 200k & 2.7472 & 4.6416 & 232.3090 & 0.8239 & 0.5590  \\
    Rectified Flow & 400k & 2.1572 & 4.5022 & 261.1203 & 0.8210 & 0.5871  \\

    \bottomrule
    \end{tabular}
    \caption{\textbf{Proof-of-concept experiments of class-conditional generation on ImageNet 256$\times$256.} Rectified flow achieves faster convergency compared to DDPM.}\label{tab:toy-fid}
\end{table}

\begin{itemize}[align=parleft, left=0pt, labelsep=0.5em]
    \item \textbf{Stage-1: Text-Semantic Pairing.} In the initial stage, we focus on establishing a solid understanding of text-to-image relationships by pretraining \ours on text-to-image tasks. This step is critical for grounding the model in basic semantic comprehension, enabling it to learn to associate textual prompts with high-level visual semantics. Through this process, the model develops a reliable capacity for representing visual concepts essential for both image and video generation, such as object attributes, spatial configurations, and contextual coherence.

    \item \textbf{Stage-2: Image-and-video joint learning.} Building on the foundational capabilities of text-to-semantic pairing, we extend \ours to joint learning across both image and video data. This stage leverages the unified framework of \ours, which employs a global attention mechanism adaptable to both images and videos. Besides, acquiring a substantial volume of high-quality video data is generally more resource-intensive compared to obtaining a similar amount of high-quality image data. To address this disparity, our framework integrates images and videos into unified token sequences during training, enabling the rich information inherent in high-quality images to enhance the generation of video frames~\citep{chen2024gentron}. By curating a carefully balanced dataset of images and videos, \ours not only gains the capability to generate both high-quality images and videos but also enhances the visual quality of videos by leveraging the rich information from high-quality image data.

    \item \textbf{Stage-3: Modality-specific finetuning.} In the final stage, we fine-tune \ours for each specific modality to further enhance its output quality. For text-to-image generation, we implement image-centric adjustments aimed at producing more visually compelling images. For text-to-video generation, we focus on adjustments that improve temporal smoothness, motion continuity, and stability across frames, resulting in realistic and fluid video outputs.

\end{itemize}

\paragraph{Cascaded Resolution Training.} In the second stage of joint training, we adopt a cascade resolution strategy to optimize the learning process. Initially, training is conducted on low-resolution image and video data ($288\times512$), enabling the model to efficiently focus on fundamental text-semantic-motion relationships at reduced computational costs. Once these core interactions are well-established, the resolution of the training data is progressively increased, transitioning from $480\times864$ to $720\times1280$. This stepwise resolution enhancement allows \ours to refine its understanding of intricate details and improve overall image fidelity, ultimately leading to superior generation quality for both images and videos.

\subsection{Image-to-Video}
To extend \ours for adapting an \emph{image} as an additional condition for video generation, we employ a widely used strategy by using the first frame of each clip as the reference image~\citep{girdhar2023emu, blattmann2023stable, yang2024cogvideox}. The corresponding image tokens are broadcasted and concatenated with the paired noised video tokens along the channel dimension. To fully leverage the pretrained knowledge during fine-tuning, we introduce a single MLP layer for channel alignment, while preserving the rest of the model architecture identical to \ours-T2V.

\section{Infrastructure Optimization} 

To achieve scalable and efficient training of \ours, we first adopt advanced parallelism strategies~(\Cref{sec:infra-model-parallel}), to handle the challenges of long-context, large-scale models. To further optimize memory usage and balance computation with communication, we implement fine-grained Activation Checkpointing~(\Cref{sec:infra-ac}). Additionally, we integrate robust fault tolerance mechanisms from MegaScale, enabling automated fault detection and recovery with minimal disruption~(\Cref{sec:infra-robust-training}). Finally, ByteCheckpoint is utilized to ensure efficient and scalable saving and loading of training states, supporting flexibility across diverse hardware configurations~(\Cref{sec:infra-byted-ckpt}). The details of these optimizations are introduced below.

\subsection{Model Parallelism Strategies}\label{sec:infra-model-parallel}
The substantial model size and the exceptionally long sequence length (exceeding 220K tokens for the longest sequence) necessitate the adoption of multiple parallelism strategies to ensure efficient training. Specifically, we employ 3D parallelism to achieve scalability across three axes: input sequences, data, and model parameters.

\paragraph{Sequence-Parallelism (SP)}~\citep{korthikanti2023reducing, li2021sequence, jacobs2023deepspeed} slices the input across the sequence dimension for independent layers (\eg, LayerNorm) to eliminate redundant computations, reduce memory usage, and support padding for non-conforming input. We adopt \textit{Ulysses}~\citep{jacobs2023deepspeed} as our implementation, which shards samples across the sequence parallel group from the start of the training loop. During attention computation, it uses all-to-all communication to distribute query, key, and value shards, allowing each worker to process the full sequence but only a subset of attention heads. After parallel computation of attention heads, another all-to-all communication aggregates the results, recombining all heads and the sharded sequence dimension.

\paragraph{Fully Sharded Data Parallelism~(FSDP)}~\citep{pytorch_fsdp} partitions all parameters, gradients and optimizer states across the data parallel ranks. Instead of all-reduce in Distributed Data Parallelism, FSDP performs all-gather for parameters and reduce-scatter for gradients, enabling overlap with forward and backward computations to potentially reduce communication overhead. In our case, we adopt the \texttt{HYBRID\_SHARD} strategy, which combines \texttt{FULL\_SHARD} within a \emph{shard group} and parameter replication across such groups, which effectively implements data parallelism (DP). This approach minimizes communication costs by limiting all-gather and reduce-scatter operations.

\subsection{Activation Checkpointing}\label{sec:infra-ac}

While the parallelism methods discussed in \Cref{sec:infra-model-parallel} provide significant memory savings and enable large-scaling training with long sequences, they inevitably introduce communication overhead among ranks, which can lead to suboptimal overall performance. To address this issue and better balance the computation and communication by maximizing their overlap in the profiling trace, we designed a fine-grained Activation Checkpointing~(AC)~\citep{chen2016training} strategy. Specifically, we implemented selective activation checkpointing to minimize the number of layers requiring activation storage while maximizing GPU utilization.

\subsection{Cluster Fault Tolerance}\label{sec:infra-robust-training}

Scaling \ours training to large-scale GPU clusters inevitably introduces fault scenarios, which can reduce training efficiency. The likelihood of encountering failures increases with the number of nodes, as larger systems have a higher probability of at least one node failing. These disruptions can extend training time and increase costs. To enhance stability and efficiency at scale, we adopted fault tolerance techniques from MegaScale~\citep{jiang2024megascale}, including self-check diagnostics, multi-level monitoring, and fast restart/recovery mechanisms. These strategies effectively mitigate the impact of interruptions, enabling \ours to maintain robust performance in large-scale generative modeling tasks.

\subsection{Saving and Loading Training Stages}\label{sec:infra-byted-ckpt}
Checkpointing training states—such as model parameters, exponential moving average (EMA) parameters, optimizer states, and random states—is crucial for training large-scale models, particularly given the increased likelihood of cluster faults. Reloading checkpointed states ensures reproducibility, which is essential for model reliability and debugging potential issues, including those caused by unintentional errors or malicious attacks.

To support scalable large-scale training, we adopt ByteCheckpoint~\citep{wan2024bytecheckpoint} as our checkpointing solution. It not only enables parallel saving and loading of partitioned checkpoints with high I/O efficiency but also supports resharding distributed checkpoints. This flexibility allows seamless switching between different training scales, accommodating varying numbers of ranks and diverse storage backends. 
In our setup, checkpointing an 8B model across over thousands of GPUs blocks training for less than 4 seconds, which is negligible compared to the overall forward and backward computation time per iteration.

\section{Data Curation Pipeline}

\begin{figure}[t]
    \centering
    \includegraphics[width=0.98\linewidth]{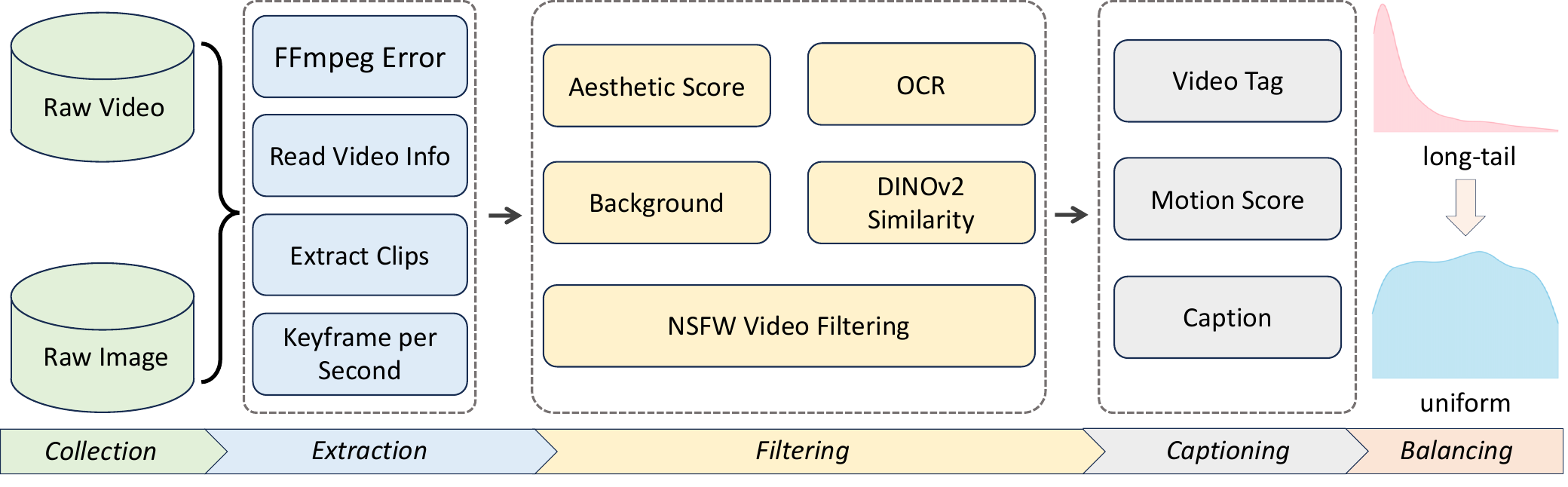}
    \caption{\textbf{The data curation pipeline in \ours.} Given a large volume of video/image data collected from Internet, we generate high-quality video/image-text pairs through a series of data filtering, captioning and balancing steps.}
    \label{fig:data-pipeline}
\end{figure}

We unblock the data volume that is utilized for industry-grade video/image generation models. Our data curation pipeline, illustrated in \Cref{fig:data-pipeline}, consists of five main stages: (1) image and video collection, (2) video extraction and clipping, (3) image and video filtering, (4) captioning, and (5) data distribution balancing. We describe the details of data curation procedure below.

\subsection{Data Overview}
We collet raw image and video data from a variety of sources, including publicly available academic datasets, internet resources, and proprietary datasets obtained through partnerships with collaborating organizations. After rigorous filtering, the final training dataset for \ours consists of approximately 160M image-text pairs and 36M video-text pairs, encompassing both publicly available datasets and internally curated proprietary datasets. The detailed composition of these resources is outlined as follows:
\begin{itemize}[align=parleft, left=0pt, labelsep=0.5em]
    \item \textbf{Text-to-Image Data.} Our text-to-image training dataset includes 100M public samples from LAION~\citep{schuhmann2022laion} and 60M high-quality, internal samples. We use public data for pre-training and internal data for fine-tuning. 
    \item \textbf{Text-to-Video Data. } Our T2V training dataset includes 11M public clips and 25M in-house clips. The former include Panda-70M~\citep{chen2024panda}, InternVid~\citep{wang2023internvid}, OpenVid-1M~\citep{nan2024openvid}, and Pexels~\citep{pku_yuan_lab_and_tuzhan_ai_etc_2024_10948109}. Rather than directly using these datasets, we apply a data curation pipeline to keep high-quality samples. 
\end{itemize}

\subsection{Data Processing and Filtering}
To construct a high-quality video dataset, we implement a comprehensive processing pipeline comprising several key stages. Raw videos are first preprocessed and standardized to address inconsistencies in encoding formats, durations, and frame rates. Next, a two-stage video clipping method segments videos into meaningful and diverse clips of consistent length. Additional filtering processes are applied, including visual aesthetic filtering to retain photorealistic and visually rich clips, OCR filtering to exclude videos with excessive text, and motion filtering to ensure balanced motion dynamics. In addition, the multi-level training data is segmented based on resolution and corresponding filtering thresholds for DINO similarity, aesthetic score, OCR text coverage, and motion score, as summarized in \Cref{tab:multi_stage_data}. We provide the details of each processing step as follows.

\Cref{tab:video-basic-parameters} presents the key parameters and their corresponding thresholds used for video quality assessment. Each parameter is essential in ensuring the generation and evaluation of high-quality videos. The Duration parameter specifies that raw video lengths should be at least 4 seconds to capture meaningful temporal dynamics. The Resolution criterion ensures that the minimum dimension (either height or width) of the video is no less than 480 pixels, maintaining adequate visual clarity. The Bitrate, which determines the amount of data processed per second during playback, requires a minimum of 500 kbps to ensure sufficient quality, clarity, and manageable file size. Videos with low bitrate typically correspond to content with low complexity, such as static videos or those featuring pure color backgrounds. Finally, the Frame Rate enforces a standard of at least 24 frames per second (film standard) or 23.976 frames per second (NTSC standard) to guarantee smooth motion and prevent visual artifacts. These thresholds collectively establish a baseline for evaluating and generating high-quality video content.

\begin{table}[t]
    \centering
    \begin{tabular}{m{0.15\linewidth} m{0.4\linewidth} m{0.35\linewidth}}
        \toprule
        Parameter & Description & Threshold  \\ \midrule
        Duration & Raw video length & $\geq$ 4 seconds  \\ \midrule
        Resolution & Width and height of the video & $min$\{ height, width\} $\geq$ 480 \\ \midrule
        Bitrate & Amount of data processed per second during playback, which impacts the video's quality, clarity, and file size & $\geq$ 500 kbps \\ \midrule
        Frame Rate & Frames displayed per second & $\geq$ 24 FPS (Film Standard) / 23.976 FPS (NTSC Standard) \\ 
        \bottomrule
        \
    \end{tabular}
    \caption{\textbf{Summary of video quality parameters and their thresholds for preprocessing.} The table outlines the criteria used to filter and standardize raw videos based on essential attributes, ensuring uniformity and compatibility in the dataset.}\label{tab:video-basic-parameters}
\end{table}

\begin{itemize}[align=parleft, left=0pt, labelsep=0.5em]
    \item \textbf{Preprocessing and Standardization of Raw Videos.} Videos collected from the internet often require extensive preprocessing to address variations in encoding formats, durations, and frame rates. Initially, we perform a primary filtering step based on fundamental video attributes such as duration, resolution, bitrate. The specific filtering criteria and corresponding thresholds are detailed in \Cref{tab:video-basic-parameters}. This initial filtering step is computationally efficient compared to more advanced, model-based filtering approaches, such as aesthetic~\citep{schuhmann2022laion} evaluation models. Following this stage, the raw videos are standardized to a consistent coding format, H.264~\citep{wiegand2003overview}, ensuring uniformity across the dataset and facilitating subsequent processing stages.
    
    \item \textbf{Video Clips Extraction.}
     We employ a two-stage video clipping method for this stage. First, we use PySceneDetect~\citep{Castellano_PySceneDetect} for shot boundary detection, resulting coarse-grained video clips from raw videos. Next, we further refine the video clips by sampling one frame per second, generating DINOv2~\citep{oquab2023dinov2} features and calculating cosine similarity between adjacent frames. When similarity falls below a set threshold, we mark a shot change and further divide the clip. Specifically, as shown in \Cref{tab:multi_stage_data}, for video resolutions around $480 \times 864$, we segmented the video clips where the DINO similarity between adjacent frames exceeds 0.85. For resolutions greater than $720 \times 1280$, the threshold is set at 0.9.
    Besides, to standardize length, we limit clips to a maximum of 10 seconds. Furthermore, we consider the similarity between different clips derived from the same source video to ensure diversity and maintain quality. Specifically, we compute the perceptual hashing~\citep{imagehash} values of keyframes from each clip and compare them. If two clips have similar hash values, indicating significant overlap, we retain the clip with a higher aesthetic score. This ensures that the final dataset includes diverse and high-quality video clips.
    
    \item \textbf{Visual Aesthetic Filtering.} To assess the visual quality of the videos, we utilize aesthetic models~\citep{schuhmann2022laion} to evaluate the keyframes. The aesthetic scores of the keyframes are averaged to obtain an overall aesthetic score for each video. For videos with resolutions around $480 \times 864$, those with an aesthetic score below 4.3 are discarded, while for resolutions exceeding $720 \times 1280$, the threshold is raised to 4.5. This filtering process ensures that the selected clips are photorealistic, visually rich, and of high aesthetic quality.
    
    \item \textbf{OCR Filtering.} To exclude videos with excessive text, we employ an internal OCR model to detect text within the keyframes. The OCR model identifies text regions, and we calculate the text coverage ratio by dividing the area of the largest bounding box detected by the total area of the keyframe. Videos with a text coverage ratio exceeding predefined thresholds are discarded. Specifically, for videos with resolutions around $480 \times 864$, the threshold is set at 0.02, while for resolutions exceeding $720 \times 1280$, the threshold is reduced to 0.01. This process effectively filters out videos with excessive text content.
    
    \item \textbf{Motion Filtering.} Unlike images, videos require additional filtering based on motion characteristics. To achieve this, we utilize RAFT~\citep{teed2020raft} to compute the mean optical flow of video clips, which is then used to derive a motion score. For videos with resolutions around $480 \times 864$, clips with motion scores below 0.3 (indicating low motion) or above 20.0 (indicating excessive motion) are excluded. For resolutions exceeding $720 \times 1280$, the thresholds are adjusted to 0.5 and 15.0, respectively. Furthermore, to enhance motion control, the motion score is appended to each caption.
\end{itemize}

\begin{table}[t]
\centering
\tablestyle{5pt}{1.12}
\begin{tabular}{ccccccc}
\toprule
Stage & Amount & Resolution & DINO-Sim.  & Aesthetic & OCR & Motion \\
\midrule
480p & 36M & $\geq$ 480$\times$864 &$\geq$0.85  &$\geq$ 4.3 & <= 0.02 &  0.3 $\leq$ score $\leq$ 20.0 \\
720p & 24M & $\geq$ 720$\times$1280  &$\geq$0.90   &$\geq$ 4.5 & <= 0.01 & 0.5 $\leq$ score $\leq$ 15.0 \\
1080p & 7M &$\geq$ 1080$\times$1920  &$\geq$0.90   &$\geq$ 4.5 & <= 0.01  & 0.5 $\leq$ score $\leq$ 8.0  \\
\bottomrule
\end{tabular}
\caption{\textbf{Overview of multi-stage training data}.This table summarizes the thresholds for each filtering criterion, including resolution, DINO similarity, aesthetic score, OCR text coverage, motion score, and the corresponding data quantities.}\label{tab:multi_stage_data}
\end{table}

\subsection{Captioning}\label{sec:captions}
Detailed captions are essential for enabling the model to generate text-aligned images/videos precisely. For images, we use InternVL2.0~\citep{chen2024far} to generate dense captions for each sample. To caption video clips, we start with InternVL2.0~\citep{chen2024far} for keyframe captions, followed by Tarsier2~\citep{yuan2025tarsier2} for video-wide captions. Note that the Tarsier2 model can inherently describe camera motion types (\eg, \emph{zoom in}, \emph{pan right}) in videos, eliminating the need for a separate prediction model and simplifying the overall pipeline compared to previous work such as \citep{polyak2024movie}.
Next, we utilize Qwen2~\citep{yang2024qwen2} to merge the keyframe and video captions. Besides, we also empirically found that adding the motion score (calculated by RAFT~\citep{teed2020raft}) to the captions improves motion control for video generation. This approach enables users to specify different motion scores in prompts to guide the model in generating videos with varied motion dynamics.

\begin{figure}[t]
    \centering
    \begin{subfigure}{\linewidth}
        \centering
        \includegraphics[width=0.95\linewidth]{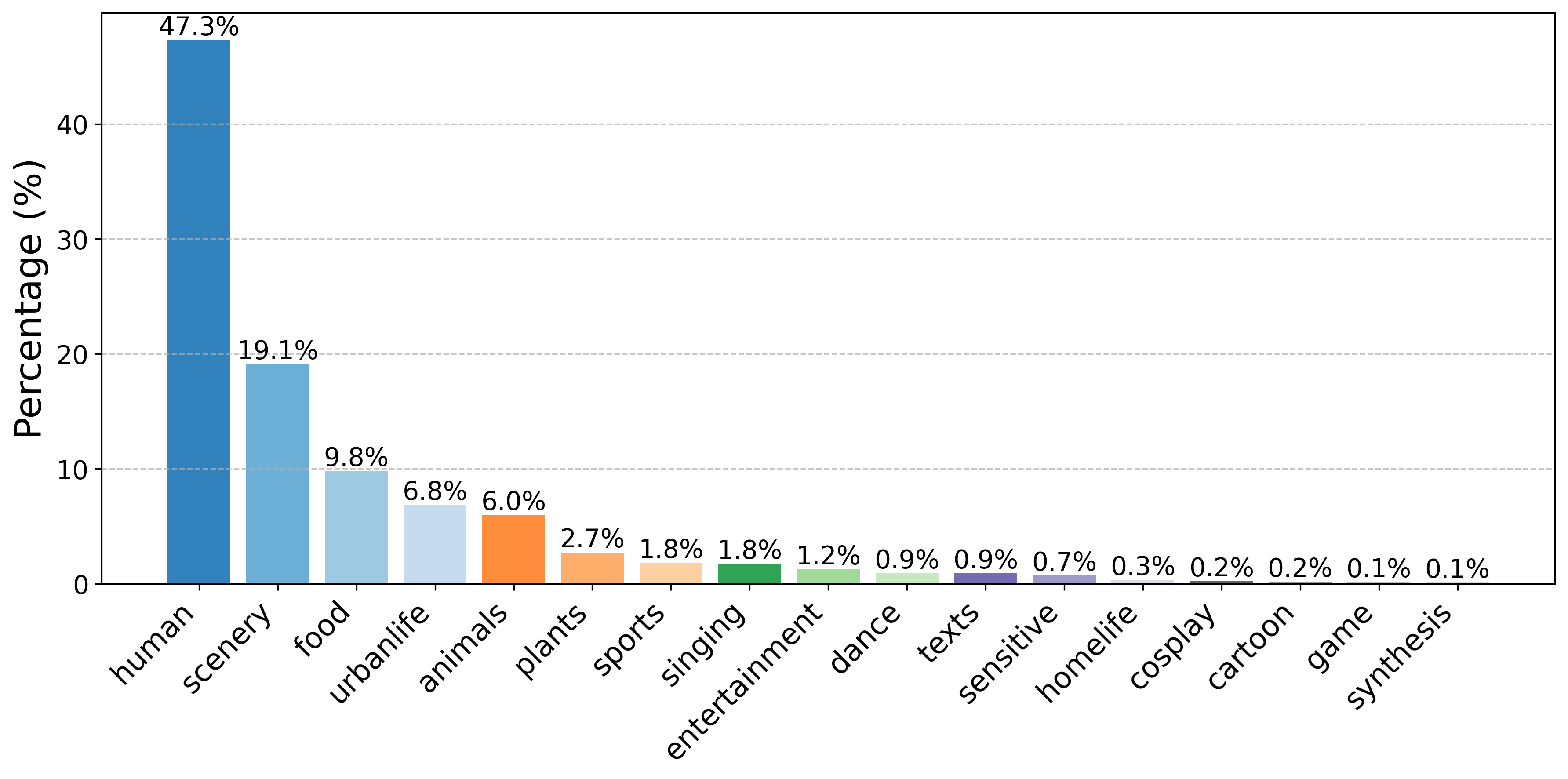} 
        \caption{Semantic distribution of video clips.}
        \label{fig:data-semantic-distribution}
    \end{subfigure}
    \vspace{5pt}
    \begin{subfigure}{\linewidth}
        \centering
        \includegraphics[width=0.95\linewidth]{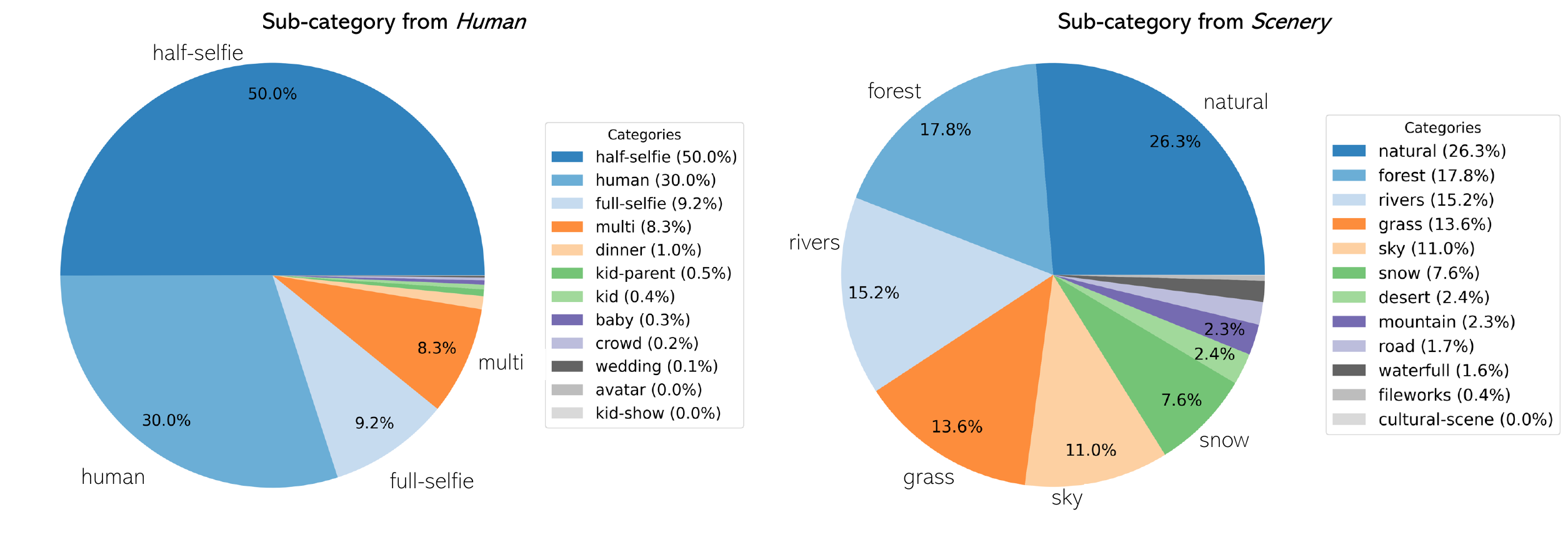} 
        \caption{The balanced semantic distribution of subcategories.}\label{fig:data-balanced}
    \end{subfigure}
    \caption{\textbf{Training data distributions.} The balanced semantic distribution of primary categories and subcategories are shown in (a) and (b), respectively.}
    \label{fig:video-data-distribution}
\end{figure}

\subsection{Training Data Balancing}\label{sec:balance}
The model’s performance are significantly influenced by the data distribution, especially for video data. To balance the video training data, we first use an internal video classification model to generate semantic tags for the videos. We then adjust the data distribution based on these semantic tags to ensure a balanced representation across categories.

\begin{itemize}[align=parleft, left=0pt, labelsep=0.5em]
    \item \textbf{Data Semantic Distribution.} The video classification model assigns a semantic tag to each video based on four evenly sampled keyframes. The model categorizes videos into 9 primary classes (\eg, human, scenery, animals, food) and 86 subcategories (\eg, half-selfie, kid, dinner, wedding). \Cref{fig:data-semantic-distribution} presents the semantic distribution across our filtered training clips, with humans, scenery, food, urban life, and animals as the predominant categories.
    \item \textbf{Data Balancing.} The quality of the generated videos is closely tied to the semantic distribution of the training data. Videos involving humans pose greater modeling challenges due to the extensive diversity in appearances, whereas animals and landscapes exhibit more visual consistency and are relatively easier to model. To address this disparity, we implement a data-balancing strategy that emphasizes human-related content while ensuring equitable representation across subcategories within each primary category. Overrepresented subcategories are selectively down-sampled, whereas underrepresented ones are augmented through artificial data generation and oversampling techniques. Balanced data distribution is shown in \Cref{fig:data-balanced}.
\end{itemize}

\section{Experiments}

\subsection{Text-to-Image Results}

\begin{table}[t]
    \centering
    \tablestyle{1.5pt}{1.12}
        \begin{tabular}{l|c|c|ccc|c}
            \toprule
            \multirow{2}{*}{Method}&   & \multicolumn{1}{c|}{GenEval}  &  \multicolumn{3}{c|}{T2I-CompBench}  &  \multicolumn{1}{c}{DPG-Bench}  \\
             & Text Enc. & Overall &  Color  & Shape & Texture &  Average   \\
            \midrule
            SDv1.5~\citep{rombach2022high}       &  CLIP ViT-L/14     & 0.43 & 0.3730 & 0.3646 & 0.4219 & 63.18   \\
            DALL-E 2~\citep{ramesh2022hierarchical} & CLIP ViT-H/16  & 0.52 & 0.5750 & 0.5464 & 0.6374 & -   \\
            SDv2.1~\citep{rombach2022high}          &  CLIP ViT-H/14  & 0.50 & 0.5694 & 0.4495 &  0.4982 & -   \\
            SDX~\citep{podell2023sdxl}             &  CLIP ViT-bigG  & 0.55 & 0.6369 & 0.5408 & 0.5637 & 74.65 \\
            PixArt-$\alpha$~\citep{chen2023pixart} &  Flan-T5-XXL  & 0.48 & 0.6886 & 0.5582 & 0.7044 & 71.11   \\
            DALL-E 3~\citep{betker2023improving}       & Flan-T5-XXL   &~~0.67$^\dagger$  & 0.8110$^\dagger$ & 0.6750$^\dagger$ & 0.8070$^\dagger$ & 83.50$^\dagger$    \\ 
            GenTron~\citep{chen2024gentron} & CLIP T5XXL &  -  & 0.7674 & 0.5700 & 0.7150  & - \\
            SD3~\citep{esser2024scaling} &  Flan-T5-XXL & 0.74 & - & - & - & -  \\
            Show-o~\citep{xie2024show}      & Phi-1.5   & 0.53 & - & - & - & - \\
            Transfusion~\citep{zhou2024transfusion}  & -  & 0.63 & - & - & - &  - \\
            Chameleon~\citep{lu2024chameleon}   & -  & 0.39 & - & - & - & - \\
            LlamaGen~\citep{sun2024autoregressive}   & FLAN-T5 XL  & 0.32  & - & - & - & - \\
            Emu 3~\citep{wang2024emu3}             & -    &~~0.66$^\dagger$ & 0.7913$^\dagger$ & 0.5846$^\dagger$ & 0.7422$^\dagger$ & 80.60  \\
            \midrule
            \ours-T2I (2B)  & \multirow{2}{*}{FLAN-T5 XL}   & 0.70 & 0.7521 & 0.4832 & 0.6691 &  \multirow{2}{*}{83.65}  \\
            \ours-T2I (2B)$^\dagger$   &  & 0.76$^\dagger$ & 0.7561$^\dagger$ & 0.5759$^\dagger$ & 0.7071$^\dagger$ &    \\
            \bottomrule
        \end{tabular}
    \caption{\textbf{Comparison with state-of-the-art models on image generation benchmarks.} We evaluate on GenEval~\citep{ghosh2024geneval}; T2I-CompBench~\citep{huang2023t2i-compbench} and DPG-Bench~\citep{hu2024ella_dbgbench}. Following ~\citep{wang2024emu3}, we use $^\dagger$ to indicate the result with prompt rewriting.}\label{tab:text2image_evaluation}
\end{table}

 we conduct a comprehensive quantitative evaluation of \ours-T2I on widely recognized image generation benchmarks, including GenEval~\citep{ghosh2024geneval}, T2I-CompBench~\citep{huang2023t2i-compbench}, and DPG-Bench~\citep{hu2024ella_dbgbench}. Details of these benchmarks could be found in \Cref{appendix:bench-config}. The results are summarized in \Cref{tab:text2image_evaluation}.

\paragraph{Performance on GenEval.} To assess text-image alignment comprehensively, we employ the GenEval benchmark, which evaluates the correspondence between textual descriptions and visual content. Since \ours-T2I is primarily trained on dense generative captions, it exhibits a natural advantage when handling detailed prompts. To further explore this, we expand the original short prompts in GenEval with ChatGPT-4o, preserving their semantics while enhancing descriptive detail. As shown in \Cref{tab:text2image_evaluation}, \ours-T2I achieves strong performance with the original short prompts, surpassing most state-of-the-art models. With the rewritten prompts, \ours-T2I attains the highest score (0.76), demonstrating its exceptional capability in aligning detailed textual descriptions with generated images.

\paragraph{Performance on T2I-CompBench.} We further evaluate the alignment between generated images and textual conditions using the T2I-CompBench benchmark, which focuses on various object attributes such as color, shape, and texture. As illustrated in \Cref{tab:text2image_evaluation}, \ours-T2I consistently outperforms several strong baselines, including PixArt-$\alpha$~\citep{chen2023pixart}, SDXL~\citep{podell2023sdxl}, and DALL-E 2~\citep{mishkin2022dall}. Notably, the inclusion of prompt rewriting leads to improved performance across all attributes, further highlighting \ours-T2I's robustness in text-image alignment.

\paragraph{Performance on DPG-Bench.} While the aforementioned benchmarks primarily evaluate text-image alignment with short prompts, DPG-Bench is designed to test model performance on dense prompt following. This challenging benchmark includes 1,000 detailed prompts, providing a rigorous test of a model's ability to generate visually accurate outputs for complex textual inputs. As shown in the last column of \Cref{tab:text2image_evaluation}, \ours-T2I achieves the highest performance with an average score of 83.65, surpassing PixArt-$\alpha$~\citep{chen2023pixart} (71.11), DALL-E 3~\citep{betker2023improving} (83.50), and EMU3~\citep{wang2024emu3} (80.60). These results highlight \ours-T2I's superior ability to handle dense prompts and maintain high fidelity in text-image alignment.

\subsection{Text-to-Video Results}

\begin{table}[t]
\tablestyle{8pt}{1.12}
\centering
    \begin{tabular}{l|c c c}
    \toprule
    Method & Resolution & FVD ($\downarrow$) & IS ($\uparrow$ )\\
    \hline
    CogVideo (Chinese)~\citep{hong2022cogvideo} & 480$\times$480 & 751.34 & 23.55 \\
    CogVideo (English)~\citep{hong2022cogvideo} & 480$\times$480 & 701.59 & 25.27 \\
    Make-A-Video~\citep{singer2023makeavideo} &  256$\times$256 & 367.23 & 33.00 \\
    VideoLDM~\citep{blattmann2023align} & - &  550.61 & 33.45 \\
    LVDM~\citep{he2022latent}           & 256$\times$256 & 372.00 & - \\
    MagicVideo~\citep{zhou2022magicvideo} & - &  655.00 & - \\ 
    PixelDance~\citep{zeng2024make}     & -  & 242.82 & 42.10 \\
    PYOCO~\citep{ge2023preserve}        & - & 355.19 & 47.76 \\
    Emu-Video~\citep{girdhar2023emu}    & 256$\times$256 & 317.10 & 42.7 \\
    SVD~\citep{blattmann2023stable}     & 240$\times$360 & 242.02 & - \\
    \hline
    \ours-2B (ours) & 256$\times$256 & 246.17 & 45.77 $\pm$ 1.10 \\
    \ours-2B (ours) & 240$\times$360 & 254.47 & 46.64 $\pm$ 1.08 \\
    \ours-2B (ours) & 128$\times$128 & 217.24 & 42.30 $\pm$ 1.03 \\
    \bottomrule
    \end{tabular}
    \caption{\textbf{Zero-shot text-to-video performance on UCF-101}. We generate videos of different resolutions, including 256$\times$256, 240$\times$360, 128$\times$128, for comprehensive comparisons.}\label{tab:ucf101-fvd}
\end{table}

\paragraph{Performance on UCF-101.} 
We conduct experiments on UCF-101~\citep{ucf101} using zero-shot text-to-video setting. As UCF-101 only has \emph{class} labels, we utilize an video-language model, Tarsier-34B~\citep{wang2024tarsier}, to generate detailed captions for all UCF-101 videos. These captions are then used to synthesize videos with \ours. Finally, we generated 13,320 videos at different resolutions with \ours-2B model for evaluation, including  256$\times$256, 240$\times$360 and 128$\times$128. Following standard practice~\citep{skorokhodov2022stylegan}, we use the I3D model, pre-trained on Kinetics-400~\citep{carreira2017quo}, as the feature extractor. Based on the extracted features, we calculated Fr\'echet Video Distance (FVD)~\citep{unterthiner2018towards} to evaluate the fidelity of the generated videos. The results in \Cref{tab:ucf101-fvd} demonstrate that \ours consistently generates videos with lower FVD and higher IS. For instance, at a resolution of 128$\times$128, the FVD of videos generated by \ours is 217.24, achieving state-of-the-art performance and highlighting significant advantages over other methods.

\paragraph{Performance on VBench.}

As presented in \Cref{tab:vbench-mini}, we evaluate \ours-T2V against state-of-the-art models on VBench~\citep{huang2024vbench}, a comprehensive benchmark designed to assess video generation quality across 16 dimensions. \ours-T2V achieves state-of-the-art overall performance on VBench, showcasing its ability to generate high-quality videos across diverse attributes and scenarios.

Among the key metrics, \ours-T2V demonstrates notable strength in human action representation, dynamic degree, and multiple object generation, reflecting its capacity for handling complex and diverse video content. Additionally, it achieves competitive results in appearance style, quality score, and semantic alignment, highlighting its balanced performance across multiple aspects.

\begin{table}[t]
    \tablestyle{3.5pt}{1.12}
    \centering
    \begin{tabular}{lcccccccc}
   \toprule
        \multirow{2}{*}{Models} & Human  & \multirow{2}{*}{Scene} &  Dynamic  & Multiple  & Appear.  & Quality & Semantic & \multirow{2}{*}{\textbf{Overall}}  \\ 
         & Action &  & Degree & Objects & Style  & Score & Score &  \\
        \midrule
\href{https://github.com/guoyww/AnimateDiff}{AnimateDiff-V2} & 92.60 & 50.19 & 40.83 & 36.88 & 22.42 & 82.90 & 69.75 & 80.27 \\
\href{https://github.com/AILab-CVC/VideoCrafter}{VideoCrafter-2.0} & 95.00 & 55.29 & 42.50 & 40.66 & \textbf{25.13} & 82.20 & 73.42 & 80.44 \\
\href{https://huggingface.co/hpcai-tech/OpenSora-STDiT-v3}{OpenSora V1.2} & 85.80 & 42.47 & 47.22 & 58.41 & 23.89 & 80.71 & 73.30 & 79.23 \\
\href{https://github.com/showlab/Show-1}{Show-1} & 95.60 & 47.03 & 44.44 & 45.47 & 23.06 & 80.42 & 72.98 & 78.93 \\
\href{https://runwayml.com/research/introducing-gen-3-alpha}{Gen-3} & 96.40 & 54.57 & 60.14 & 53.64 & 24.31 & 84.11 & 75.17 & 82.32 \\
\href{https://pika.art}{Pika-1.0} & 86.20 & 49.83 & 47.50 & 43.08 & 22.26 & 82.92 & 71.77 & 80.69 \\
\href{https://github.com/THUDM/CogVideo}{CogVideoX-5B} & 99.40 & 53.20 & 70.97 & 62.11 & 24.91 & 82.75 & 77.04 & 81.61 \\
\href{https://klingai.kuaishou.com/}{Kling} & 93.40 & 50.86 & 46.94 & 68.05 & 19.62 & 83.39 & 75.68 & 81.85 \\
\href{https://github.com/mira-space/Mira}{Mira} & 63.80 & 16.34 & 60.33 & 12.52 & 21.89 & 78.78 & 44.21 & 71.87 \\
\href{https://causvid.github.io/}{CausVid} & \textbf{99.80} & 56.58 & \textbf{92.69} & 72.15 & 24.27 & \textbf{85.65} & 78.75 & 84.27 \\
\href{https://lumalabs.ai/dream-machine}{Luma} & 96.40 & \textbf{58.98} & 44.26 & \textbf{82.63} & 24.66 & 83.47 & \textbf{84.17} & 83.61 \\
\href{https://github.com/Tencent/HunyuanVideo}{HunyuanVideo} & 94.40 & 53.88 & 70.83 & 68.55 & 19.80 & 85.09 & 75.82 & 83.24 \\
\midrule
\textbf{Goku} (ours) & 97.60 & 57.08 & 76.11 & 79.48 & 23.08 & 85.60 & 81.87 & \textbf{84.85} \\
\bottomrule
    \end{tabular}
    \caption{\textbf{Comparison with leading T2V models on VBench.} \ours achieves state-of-the-art overall performance. Detailed results across all 16 evaluation dimensions are provided in \Cref{tab:vbench-full} in the Appendix.}\label{tab:vbench-mini}
\end{table}

For detailed results on all 16 evaluation dimensions, we refer readers to \Cref{tab:vbench-full} in the Appendix. This comprehensive analysis underscores \ours-T2V's superiority in video generation compared to prior approaches.

\subsection{Image-to-Video}

\begin{figure}[ht]
\centering
\includegraphics[width=0.96\linewidth]{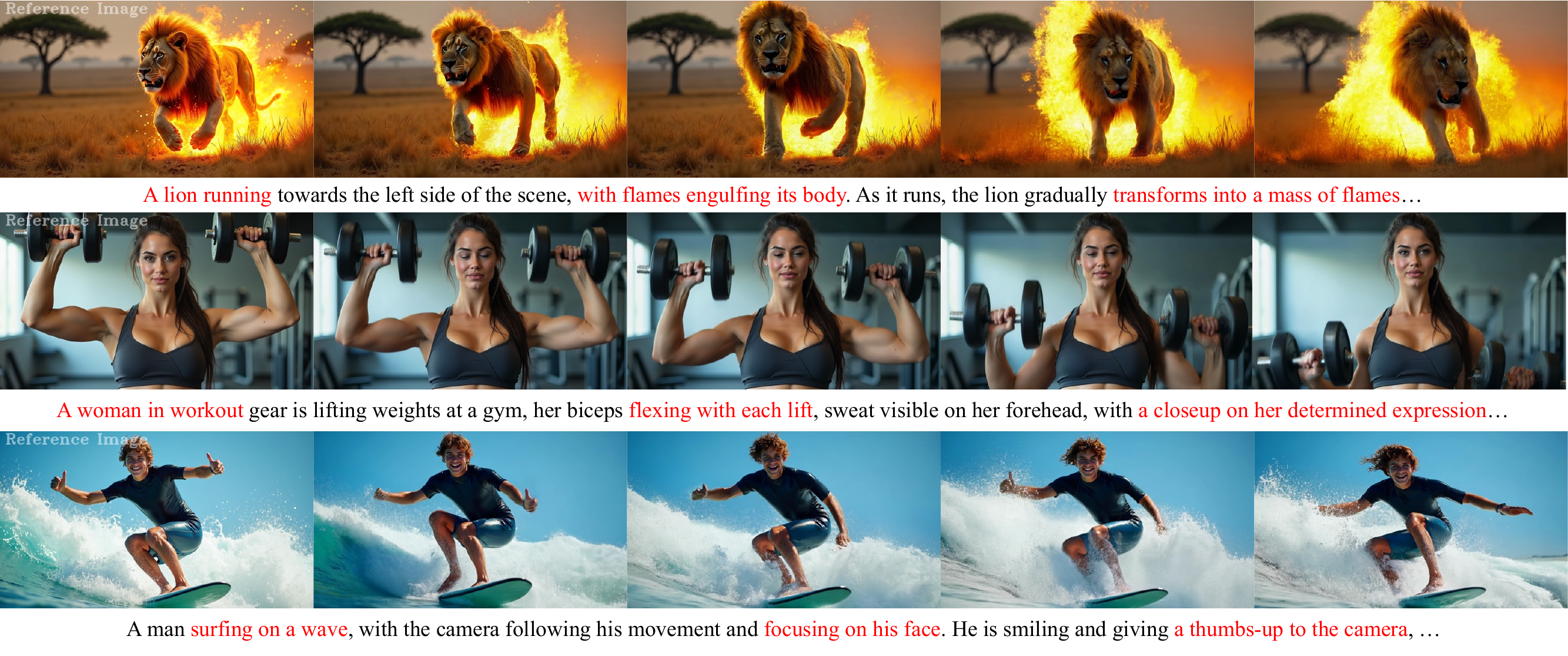} \\
\caption{\textbf{Samples of \ours-I2V.} Reference images are presented in the leftmost columns. We omitted redundant information from the long prompts, displaying only the key details in each one. Key words are highlighted in \textcolor{red}{\textbf{RED}}.}\label{fig:i2v}
\end{figure}

\begin{figure}[ht]
    \centering
    \begin{subfigure}[b]{\linewidth}
        \centering
        \includegraphics[width=0.7\linewidth]{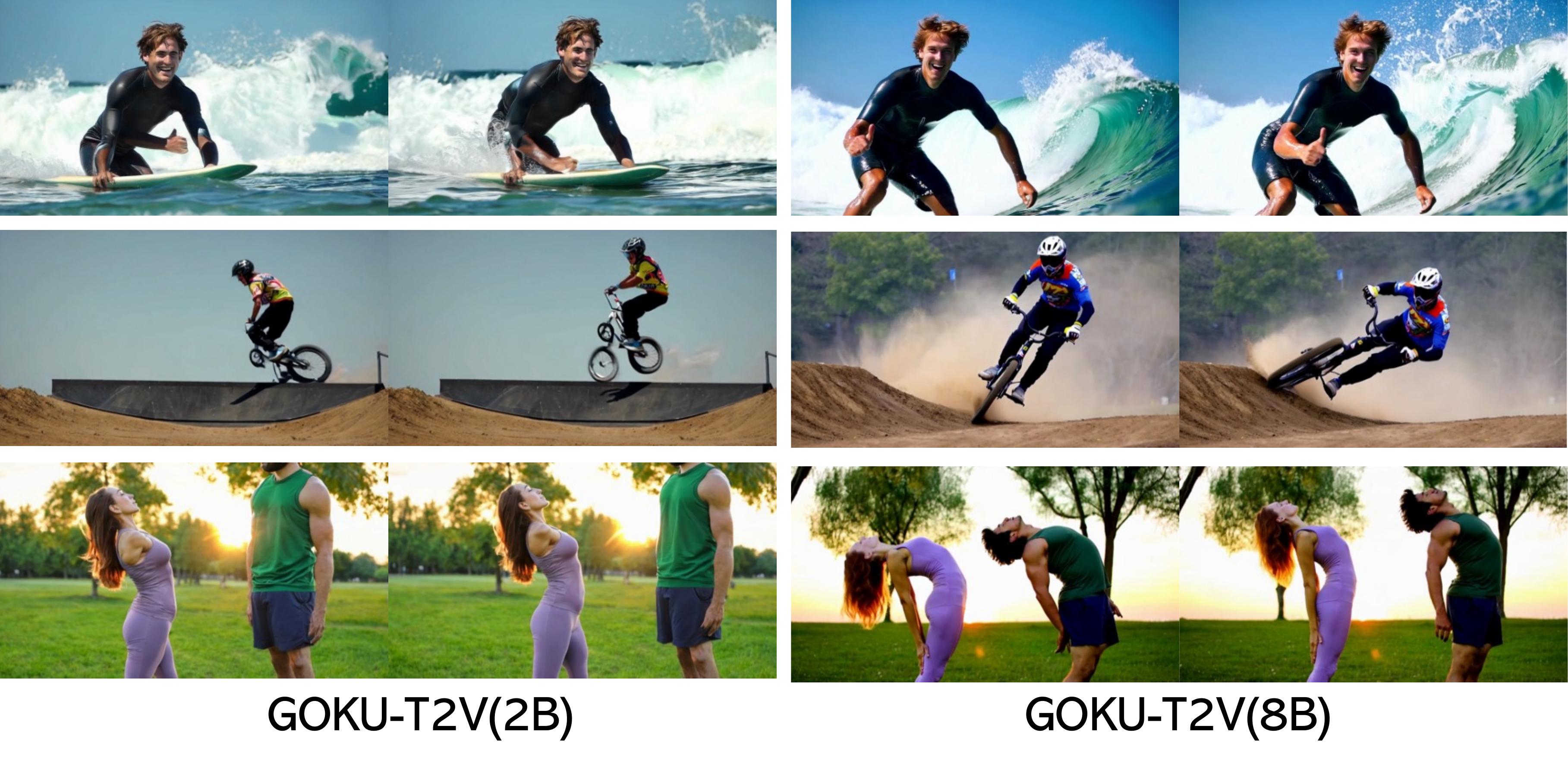} 
        \caption{Model Scaling}
        \label{fig:ablation_1}
    \end{subfigure}
    \vfill
    \begin{subfigure}[b]{\linewidth}
        \centering
    \includegraphics[width=0.7\linewidth]{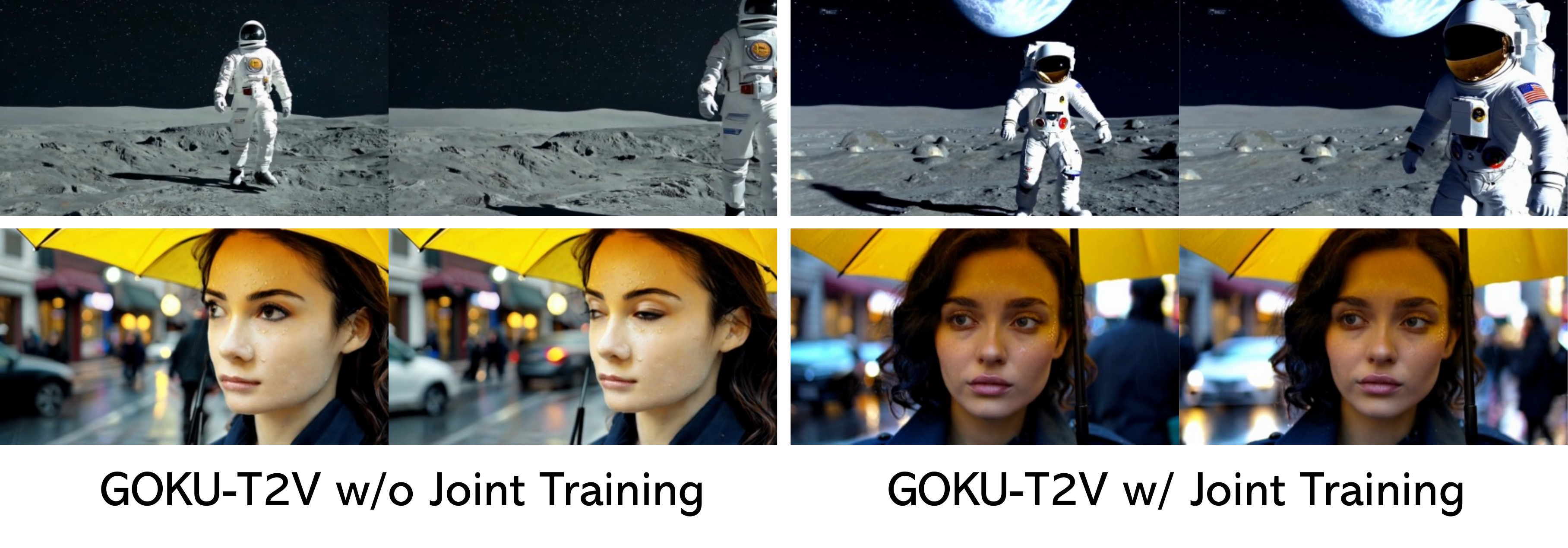} 
        \caption{Joint Training}
        \label{fig:ablation_2}
    \end{subfigure}
    \caption{\textbf{Ablation Studies of Model Scaling and Joint Training.} Fig. (a) shows the comparison between \ours-T2V(2B) and \ours-T2V(8B). Fig. (b) shows the comparison  between whether joint training is adopted or not. }
    \label{fig:video-tag}
\end{figure}

We finetune \ours-I2V from the T2V initialization with approximate 4.5M text-image-video triplets, sourced from diverse domains to ensure robust generalization. Despite the relatively small number of fine-tuning steps (10k), our model demonstrates remarkable efficiency in animating reference image while maintaining strong alignment with the accompanying text. As illustrated in \Cref{fig:i2v}, the generated videos exhibit high visual quality and temporal coherence, effectively capturing the semantic nuances described in the text. 

\begin{figure}[t]
\centering
\includegraphics[width=0.96\textwidth]{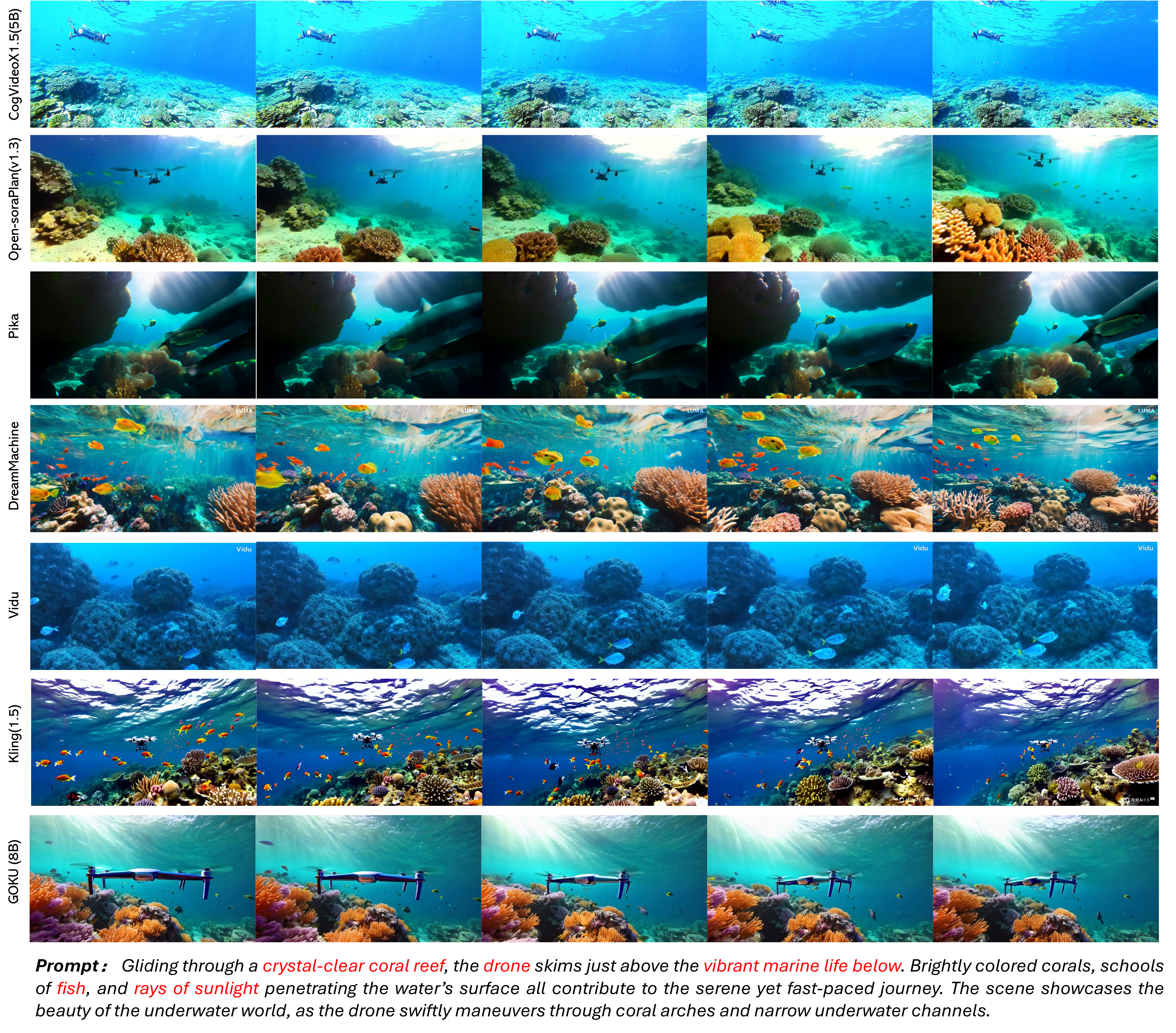} \\
\caption{\textbf{Qualitative comparisons with state-of-the-art (SoTA) video generation models.} This figure showcases comparisons with leading models, including~\citep{yang2024cogvideox}, Open-Sora Plan~\citep{pku_yuan_lab_and_tuzhan_ai_etc_2024_10948109}, Pika~\citep{pika}, DreamMachine~\citep{DreamMachine}, Vidu~\citep{bao2024vidu}, and Kling v1.5~\citep{kuaishou2024klingai}.}
\label{fig:video-sota}
\end{figure}

\subsection{Image and Video Qualitative Visualizations}\label{sec:overall-qualitative-results}

For intuitive comparisons, we conduct qualitative assessments and present sampled results in \Cref{fig:video-sota}. The evaluation includes open-source models, such as CogVideoX~\citep{yang2024cogvideox} and Open-Sora-Plan~\citep{opensora}, alongside closed-source commercial products, including DreamMachine~\citep{DreamMachine}, Pika~\citep{pika}, Vidu~\citep{bao2024vidu}, and Kling~\citep{kuaishou2024klingai}. The results reveal that some commercial models struggle to generate critical video elements when handling complex prompts. For instance, models like Pika, DreamMachine, and Vidu (rows 3–5) fail to render the skimming drone over water. While certain models succeed in generating the target drone, they often produce distorted subjects (rows 1–2) or static frames lacking motion consistency (row 6). In contrast, \ours-T2V (8B) demonstrates superior performance by accurately incorporating all details from the prompt, creating a coherent visual output with smooth motion. Additional comparisons are provided in the appendix for a more comprehensive evaluation. Furthermore, more video examples are available at \href{https://saiyan-world.github.io/goku/}{the goku homepage}.

\subsection{Ablation Studies}

\paragraph{Model Scaling.} We compared \ours-T2V models with 2B and 8B parameters. Results in \cref{fig:ablation_1} indicate that model scaling helps mitigate the generation of distorted object structures, such as the arm in \cref{fig:ablation_1} (row 1) and the wheel in \cref{fig:ablation_1} (row 2). This aligns with findings observed in large multi-modality models.

\paragraph{Joint Training.} We further examine the impact of joint image-and-video training. Starting from the same pretrained \ours-T2I (8B) weights, we fine-tuned \ours-T2V (8B) on 480p videos for an equal number of training steps, with and without joint image-and-video training. As shown in \cref{fig:ablation_2}, \ours-T2V without joint training tends to generate low-quality video frames, while the model with joint training more consistently produces photorealistic frames.

\section{Conclusion}
In this work, we presented \ours, a novel model for joint image-and-video generation for industry-standard performance. Through an advanced data curation process and a robust model architecture, \ours delivers high-quality outputs by ensuring both fine-grained data selection and effective integration of image and video modalities. Key components, such as the image-video joint VAE and the application of rectified flow, facilitate seamless token interaction across modalities, establishing a shared latent space that enhances model adaptability and attention across tokens. Empirical results highlight \ours's superiority in commercial-grade visual generation quality.

\subsection*{Acknowledgements}
We sincerely appreciate the support of our collaborators at ByteDance who contributed to this work. Xibin Wu, Chongxi Wang, Yina Tang, Fangzhou Ai, Yi Ren, Wei Wang, Chen Chen, Colin Young, Bobo Zeng, Ge Bai, Yi Fu, Ruoyu Guo, Prasanna Raghav, Weiguo Feng, Xugang Ye, Adithya Sampath, Aaron Shen, Da Tang, Yuan Fang, Qijun Gan, Chen Zhang, Zhenhui Ye, Pan Xie, Houmin Wei, Gaohong Liu, Zherui Liu, Chenyuan Wang, Yun Zhang, Kaihua Jiang, Zhuo Jiang, Yang Bai, Weiqiang Lou, Hongkai Li, Xi Yang, Shuguang Wang, Junru Zheng, Zuquan Song, Zixian Du, Jingzhe Tang, Yongqiang Zhang, Mingji Han, Heng Zhang, Li Han,  Sophie Xie, Shuo Li, Xinzhi Yao, Peng Li, Lianke Qin, Dongyang Wang, Yang Cheng, Chundian Liu, Wenhao Hao, Haibin Lin, Xin Liu 


\clearpage
\appendix

\renewcommand{\thesection}{\Alph{section}}
\renewcommand{\thesection}{Appendix \Alph{section}}

\section{Benchmark Configurations}\label{appendix:bench-config}

\paragraph{T2I-Compbench~\citep{huang2023t2i-compbench}}  We evaluate the alignment between the generated images and text conditions using T2I-Compbench, a comprehensive benchmark for assessing compositional text-to-image generation capabilities. Specifically, we report scores for color binding, shape binding, and texture binding. To evaluate these results, we employ the Disentangled BLIP-VQA model. For each attribute, we generate 10 images per prompt, with a total of 300 prompts in each category.

\paragraph{GenEval~\citep{ghosh2024geneval}} GenEval is an object-focused framework designed to evaluate compositional image properties, such as object co-occurrence, position, count, and color. For evaluation, we generate a total of 2,212 images across 553 prompts. The final score is reported as the average across tasks.

\paragraph{DPG-Bench~\citep{hu2024ella_dbgbench}} Compared to the aforementioned benchmarks, DPGBench offers longer prompts with more detailed information, making it effective for evaluating compositional generation in text-to-image models. For this evaluation, we generate a total of 4,260 images across 1,065 prompts, with the final score reported as the average across tasks.

\paragraph{VBench~\citep{huang2024vbench}} VBench is a benchmark suite for evaluating video generative models. It provides a structured Evaluation Dimension Suite that breaks down ``video generation quality" into precise dimensions for detailed assessment. Each dimension and content category includes a carefully crafted Prompt Suite and samples Generated Videos from various models.

\begin{figure}[ht]
\begin{center}
\begin{tabular}{c}
\includegraphics[width=0.82\textwidth]{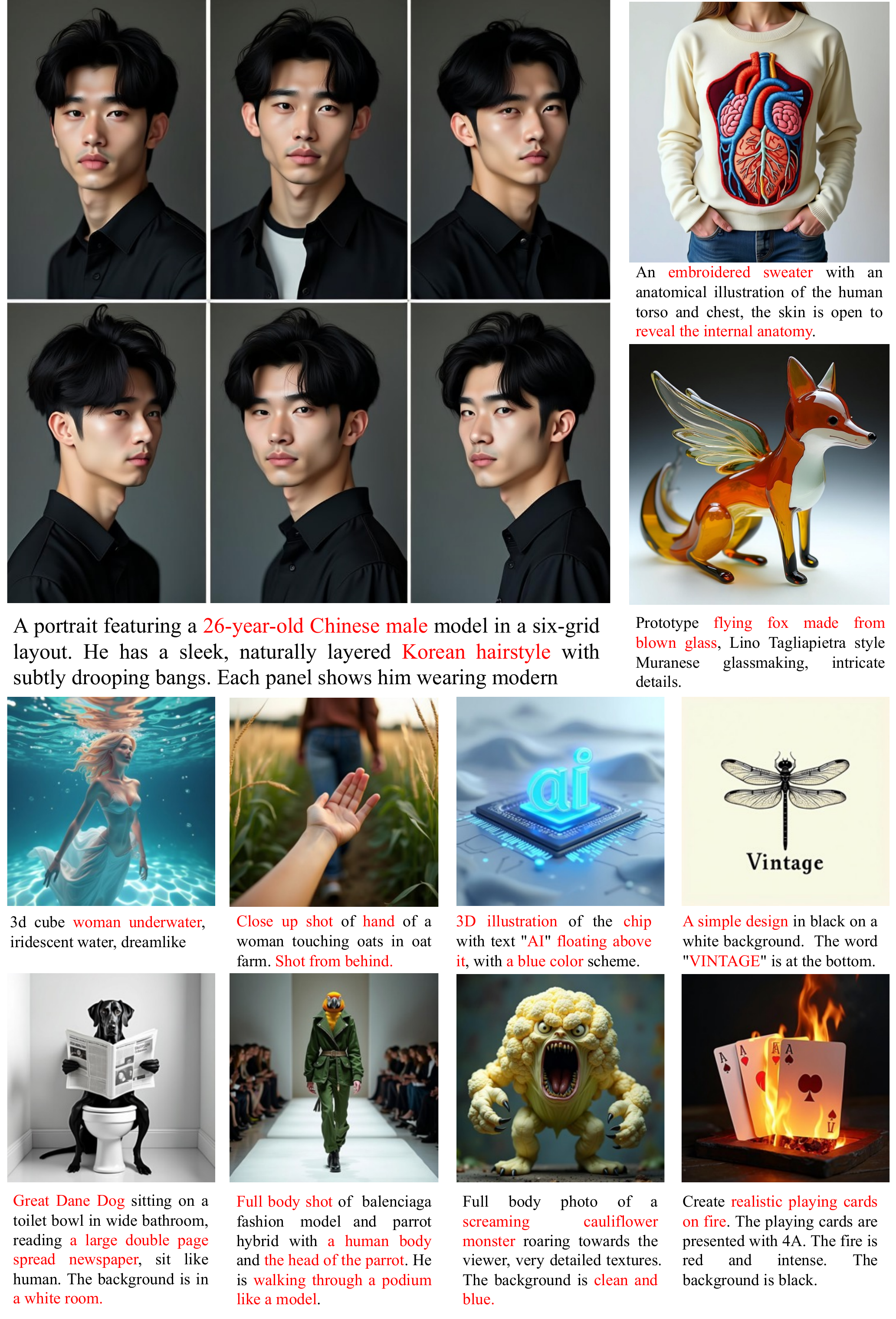} \\
\end{tabular}
\end{center}
\caption{\textbf{Qualitative samples of \ours-T2I.} Key words are highlighted in \textcolor{red}{RED}.}
\label{fig:t2i_supp}
\end{figure}

\begin{figure}[ht]
\begin{center}
\begin{tabular}{c}
\includegraphics[width=1\textwidth]{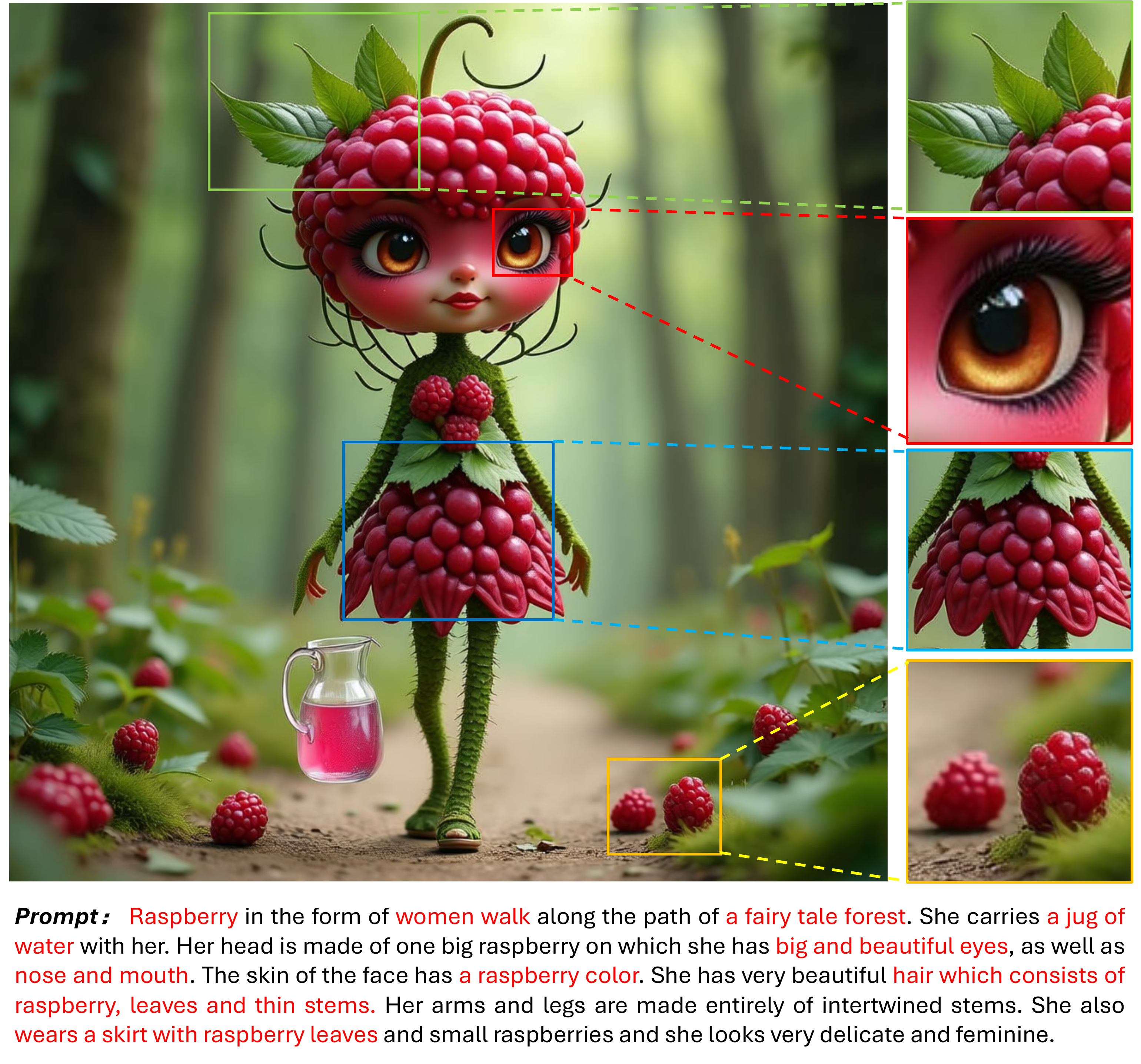} \\
\end{tabular}
\end{center}
\caption{\textbf{Qualitative samples of \ours-T2I.} Key words are highlighted in \textcolor{red}{RED}. For clarity, we zoom in on specific regions to enhance visualization.}
\label{fig:t2i_zoom_supp}
\end{figure}

\section{More Visualization Examples}
\subsection{\ours-T2I Samples Visualization}\label{sec:more-t2i-example}

We present more generated image samples with their text prompts in \Cref{fig:t2i_supp}.  The prompts are randomly selected from the Internet \footnote{\url{https://promptlibrary.org/}}.  \ours-T2I achieves strong performance in both visual quality and text-image alignment. It can interpret visual elements and their interactions from complex natural language descriptions. Notably, in \Cref{fig:t2i_zoom_supp}, \ours-T2I exhibits impressive abilities on generating images with rich details, for example, the clear textures of leaves and berries.

\subsection{\ours-T2V Samples Visualization}
\begin{figure}[ht]
\begin{center}
\begin{tabular}{c}
\includegraphics[width=1\textwidth]{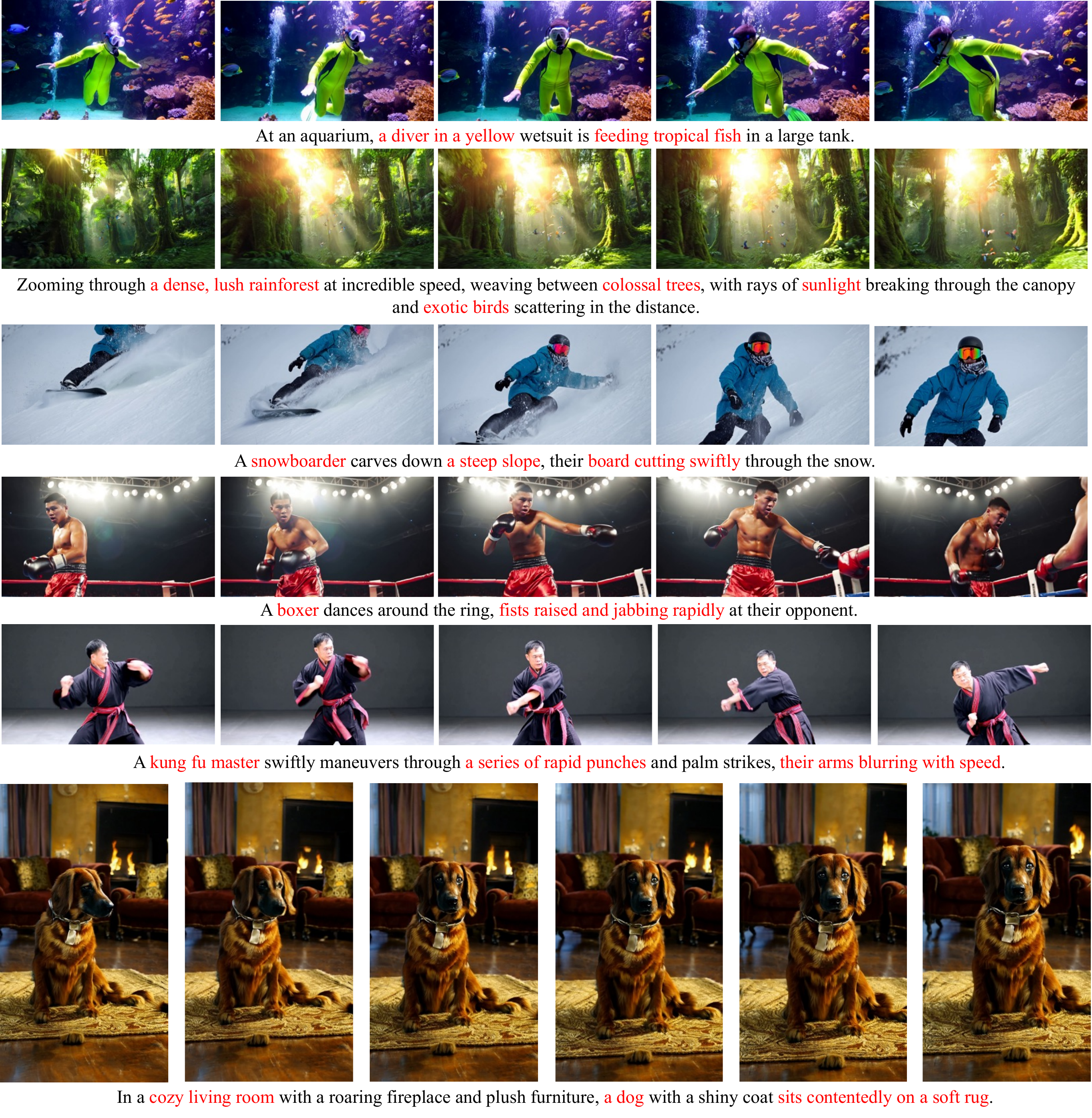} \\
\end{tabular}
\end{center}
\caption{\textbf{Qualitative samples of \ours-T2V.} Key words are highlighted in \textcolor{red}{\textbf{RED}}.}
\label{fig:t2v_supp}
\end{figure}

In \Cref{fig:t2v_supp} we show more examples generated by \ours-T2V, in both \emph{landscape} (\eg, rows one through five) and \emph{portrait} mode (\eg, the last row). \ours-T2V is capable of generating high-motion videos (\eg, skiing) and realistic scenes (\eg, forests). All videos are configured with a duration of 4 seconds, a frame rate of 24 FPS, and a resolution of 720p. For visualization, we uniformly sample five frames in temporal sequence.

\subsection{\ours-T2V Comparisons with Prior Arts}

\begin{figure}[ht]
\begin{center}
\begin{tabular}{c}
\includegraphics[width=0.9\textwidth]{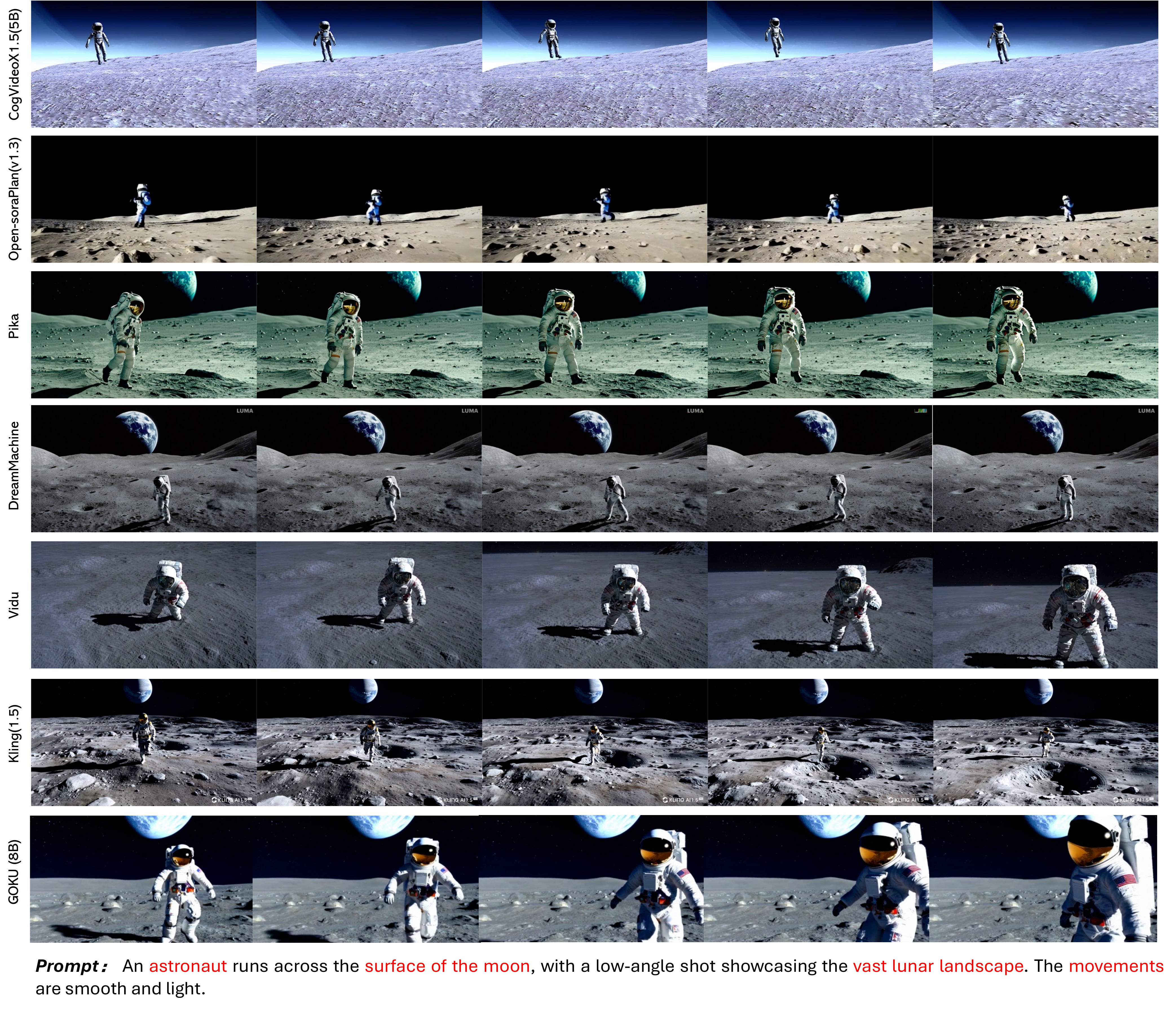} \\
\end{tabular}
\end{center}
\caption{\textbf{Qualitative comparisons of \ours-T2V with SOTA video generation methods.} Key words are highlighted in \textcolor{red}{\textbf{RED}}.}
\label{fig:video_comparisons_supp1}
\end{figure}

\begin{figure}[ht]
\begin{center}
\begin{tabular}{c}
\includegraphics[width=0.9\textwidth]{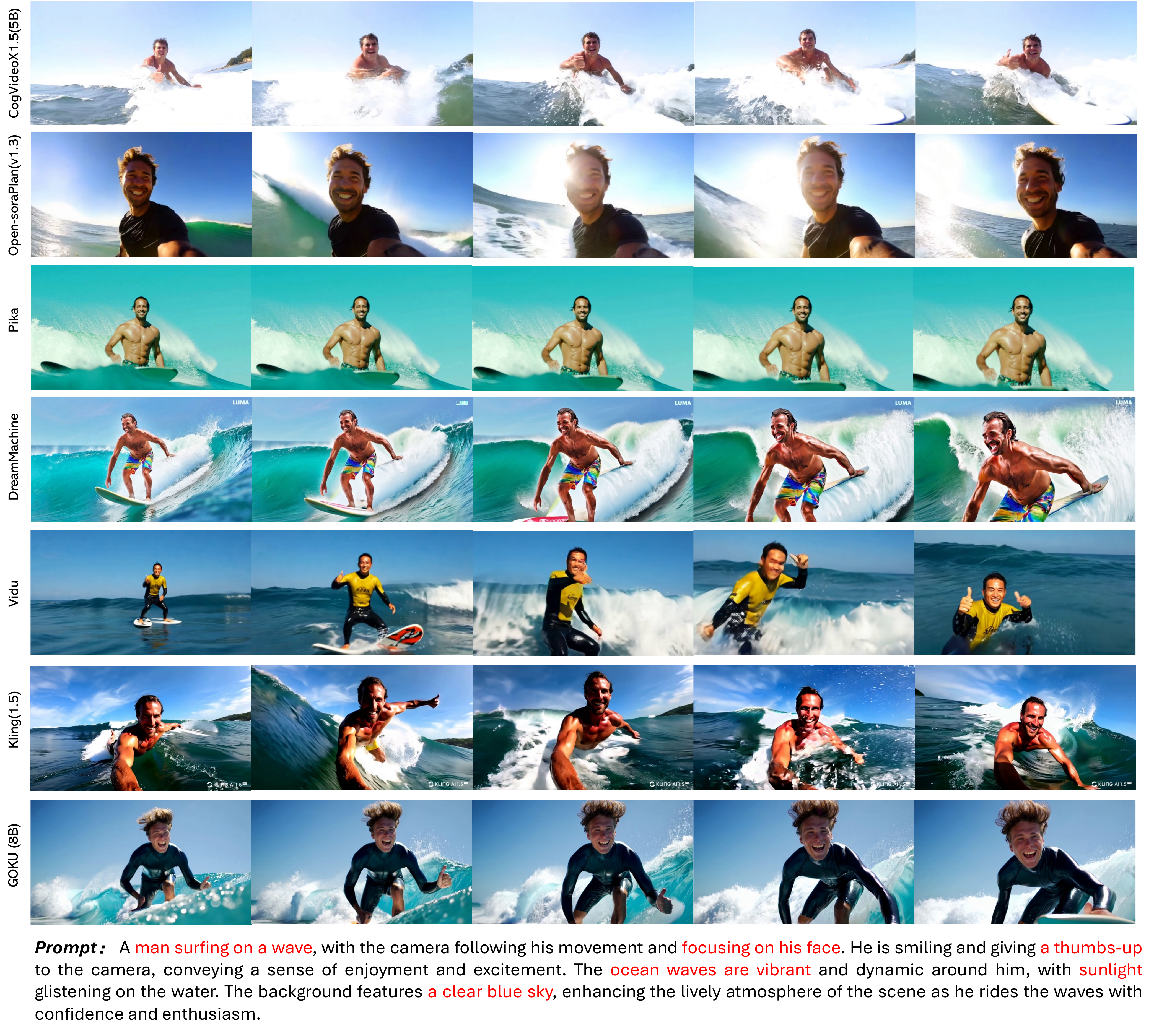} \\
\end{tabular}
\end{center}
\caption{\textbf{Qualitative comparisons of \ours-T2V with SOTA video generation methods.} Key words are highlighted in \textcolor{red}{\textbf{RED}}.}
\label{fig:video_comparisons_supp2}
\end{figure}

Additional comparisons with state-of-the-art text-to-video generation models are presented in \Cref{fig:video_comparisons_supp1} and \Cref{fig:video_comparisons_supp2}. These results demonstrate the strong performance of \ours when evaluated against both open-source models \citep{yang2024cogvideox,opensora} and commercial products \citep{pika,kuaishou2024klingai,bao2024vidu,DreamMachine}. For instance, in \Cref{fig:video_comparisons_supp2}, \ours successfully generates smooth motion and accurately incorporates the specified low-angle shot. In contrast, other models, such as CogVideoX~\citep{yang2024cogvideox}, Vidu~\citep{bao2024vidu}, and Kling~\citep{kuaishou2024klingai}, often produce incorrect objects or improper camera views.

\begin{sidewaystable}
\centering
\small
\setlength{\tabcolsep}{3pt}
\renewcommand{\arraystretch}{1}
\begin{tabular}{l|lllllllllllllllllll}
     Method  & 
    \makecell[bc]{\rotatebox{75}{Total Score}} & 
    \makecell[bc]{\rotatebox{75}{Quality Score}} & 
    \makecell[bc]{\rotatebox{75}{Semantic Score}} & 
    \makecell[bc]{\rotatebox{75}{subject consistency}} & 
    \makecell[bc]{\rotatebox{75}{background consistency}} & 
    \makecell[bc]{\rotatebox{75}{temporal flickering}} & 
    \makecell[bc]{\rotatebox{75}{motion smoothness}} & 
    \makecell[bc]{\rotatebox{75}{dynamic degree}} & 
    \makecell[bc]{\rotatebox{75}{aesthetic quality}} & 
    \makecell[bc]{\rotatebox{75}{imaging quality}} & 
    \makecell[bc]{\rotatebox{75}{object class}} & 
    \makecell[bc]{\rotatebox{75}{multiple objects}} & 
    \makecell[bc]{\rotatebox{75}{human action}} & 
    \makecell[bc]{\rotatebox{75}{color}} & 
    \makecell[bc]{\rotatebox{75}{spatial relationship}} & 
    \makecell[bc]{\rotatebox{75}{scene}} & 
    \makecell[bc]{\rotatebox{75}{appearance style}} & 
    \makecell[bc]{\rotatebox{75}{temporal style}} & 
    \makecell[bc]{\rotatebox{75}{overall consistency}} \\
\href{https://github.com/guoyww/AnimateDiff}{AnimateDiff-V2} & 80.27 & 82.90 & 69.75 & 95.30 & 97.68 & 98.75 & 97.76 & 40.83 & 67.16 & 70.10 & 90.90 & 36.88 & 92.60 & 87.47 & 34.60 & 50.19 & 22.42 & 26.03 & 27.04 \\
\href{https://github.com/AILab-CVC/VideoCrafter}{VideoCrafter-2.0} & 80.44 & 82.20 & 73.42 & 96.85 & \textbf{98.22} & 98.41 & 97.73 & 42.50 & 63.13 & 67.22 & 92.55 & 40.66 & 95.00 & \textbf{92.92} & 35.86 & 55.29 & \textbf{25.13} & 25.84 & \textbf{28.23} \\
\href{https://huggingface.co/hpcai-tech/OpenSora-STDiT-v3}{OpenSora V1.2} & 79.23 & 80.71 & 73.30 & 94.45 & 97.90 & 99.47 & 98.20 & 47.22 & 56.18 & 60.94 & 83.37 & 58.41 & 85.80 & 87.49 & 67.51 & 42.47 & 23.89 & 24.55 & 27.07 \\
\href{https://github.com/showlab/Show-1}{Show-1} & 78.93 & 80.42 & 72.98 & 95.53 & 98.02 & 99.12 & 98.24 & 44.44 & 57.35 & 58.66 & 93.07 & 45.47 & 95.60 & 86.35 & 53.50 & 47.03 & 23.06 & 25.28 & 27.46 \\
\href{https://runwayml.com/research/introducing-gen-3-alpha}{Gen-3} & 82.32 & 84.11 & 75.17 & 97.10 & 96.62 & 98.61 & 99.23 & 60.14 & 63.34 & 66.82 & 87.81 & 53.64 & 96.40 & 80.90 & 65.09 & 54.57 & 24.31 & 24.71 & 26.69 \\
\href{https://pika.art}{Pika-1.0} & 80.69 & 82.92 & 71.77 & 96.94 & 97.36 & \textbf{99.74} & \textbf{99.50} & 47.50 & 62.04 & 61.87 & 88.72 & 43.08 & 86.20 & 90.57 & 61.03 & 49.83 & 22.26 & 24.22 & 25.94 \\
\href{https://github.com/THUDM/CogVideo}{CogVideoX-5B} & 81.61 & 82.75 & 77.04 & 96.23 & 96.52 & 98.66 & 96.92 & 70.97 & 61.98 & 62.90 & 85.23 & 62.11 & 99.40 & 82.81 & 66.35 & 53.20 & 24.91 & 25.38 & 27.59 \\
\href{https://klingai.kuaishou.com/}{Kling} & 81.85 & 83.39 & 75.68 & \textbf{98.33} & 97.60 & 99.30 & 99.40 & 46.94 & 61.21 & 65.62 & 87.24 & 68.05 & 93.40 & 89.90 & 73.03 & 50.86 & 19.62 & 24.17 & 26.42 \\
\href{https://github.com/mira-space/Mira}{Mira} & 71.87 & 78.78 & 44.21 & 96.23 & 96.92 & 98.29 & 97.54 & 60.33 & 42.51 & 60.16 & 52.06 & 12.52 & 63.80 & 42.24 & 27.83 & 16.34 & 21.89 & 18.77 & 18.72 \\
\href{https://causvid.github.io/}{CausVid} & 84.27 & \textbf{85.65} & 78.75 & 97.53 & 97.19 & 96.24 & 98.05 & \textbf{92.69} & 64.15 & 68.88 & 92.99 & 72.15 & \textbf{99.80} & 80.17 & 64.65 & 56.58 & 24.27 & 25.33 & 27.51 \\
\href{https://lumalabs.ai/dream-machine}{Luma} & 83.61 & 83.47 & \textbf{84.17} & 97.33 & 97.43 & 98.64 & 99.35 & 44.26 & 65.51 & 66.55 & \textbf{94.95} & \textbf{82.63} & 96.40 & 92.33 & 83.67 & \textbf{58.98} & 24.66 & \textbf{26.29} & 28.13 \\
\href{https://github.com/Tencent/HunyuanVideo}{HunyuanVideo} & 83.24 & 85.09 & 75.82 & 97.37 & 97.76 & 99.44 & 98.99 & 70.83 & 60.36 & 67.56 & 86.10 & 68.55 & 94.40 & 91.60 & 68.68 & 53.88 & 19.80 & 23.89 & 26.44 \\
\href{here}{Goku} & \textbf{84.85} & 85.60 & 81.87 & 95.55 & 96.67 & 97.71 & 98.50 & 76.11 & \textbf{67.22} & \textbf{71.29} & 94.40 & 79.48 & 97.60 & 83.81 & \textbf{85.72} & 57.08 & 23.08 & 25.64 & 27.35 \\

\bottomrule 
\end{tabular}
\caption{\textbf{Comparison with state-of-the-art models on video generation benchmarks.} We evaluate on VBench~\citep{huang2024vbench} and compare with Gen-3~\citep{runway2023gen2}, Vchitect-2.0~\citep{Vchitect2}, VEnhancer~\citep{he2024venhancer}, Kling~\citep{kuaishou2024klingai}, LaVie-2~\citep{wang2023lavie}, CogVideoX~\citep{yang2024cogvideox}, Emu3~\citep{wang2024emu3}.}
\label{tab:vbench-full}
\end{sidewaystable}

\subsection{\ours-I2V Samples Visualization}

\begin{figure}[ht]
\begin{center}
\begin{tabular}{c}
\includegraphics[width=0.9\textwidth]{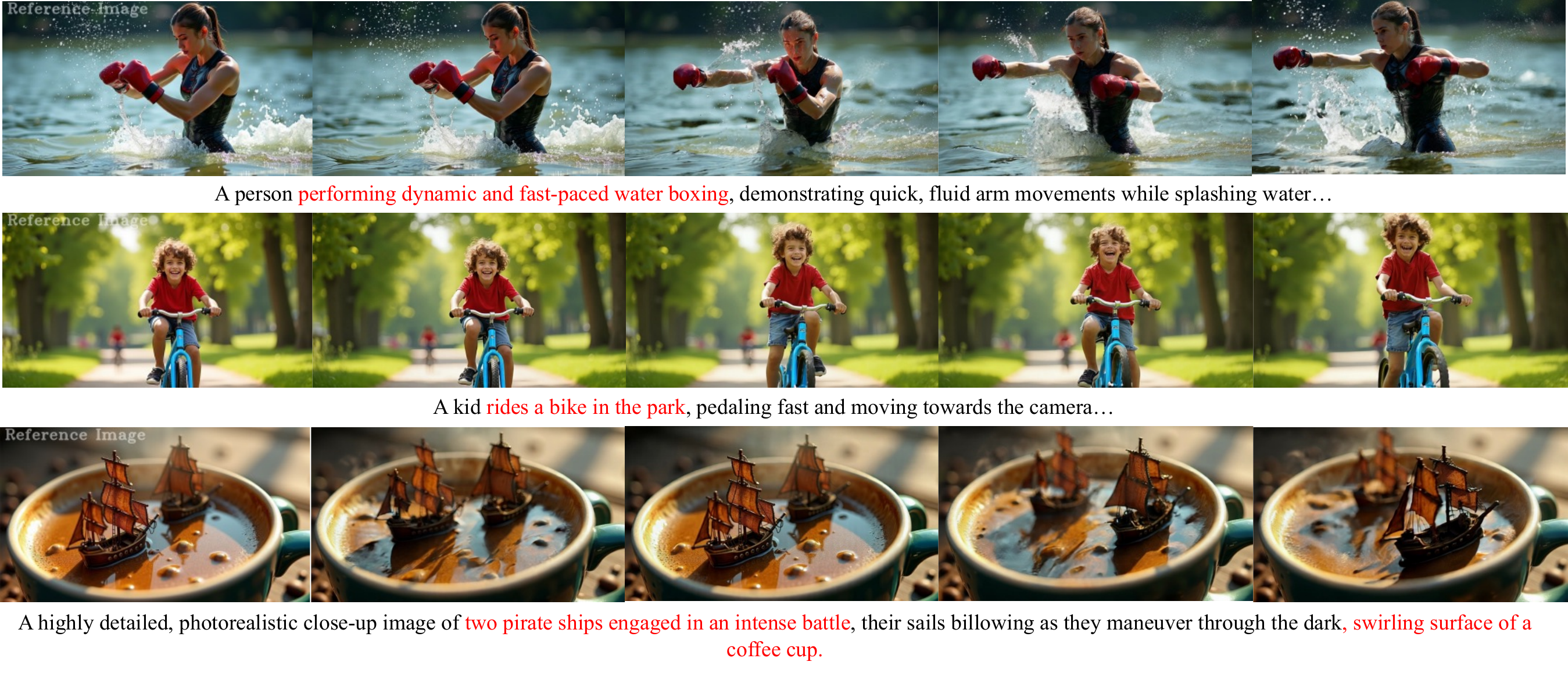} \\
\end{tabular}
\end{center}
\caption{\textbf{Qualitative samples of \ours-I2V.} Key words are highlighted in \textcolor{red}{\textbf{RED}}.}
\label{fig:i2v_supp}
\end{figure}

We present additional visualization of generated samples from \ours-I2V in \Cref{fig:i2v_supp}, which further validate the effectiveness and versatility of our approach. As shown in the figure, \ours-I2V demonstrates an impressive ability to synthesize coherent and visually compelling videos from diverse reference images, maintaining consistency in motion and scene semantics.

For instance, in the first row, the model successfully captures the dynamic and high-energy nature of water boxing, generating fluid and natural movements of splashes synchronized with the subject’s motions. In the second row, the sequence of a child riding a bike through a park illustrates the model’s proficiency in creating smooth and realistic forward motion while preserving environmental consistency. Finally, the third row showcases the model’s ability to handle creative and imaginative scenarios, as seen in the detailed depiction of pirate ships battling atop a swirling coffee cup. The photorealistic rendering and accurate motion trajectories underscore the model’s robustness in both realism and creativity.

These examples highlight \ours-I2V’s capacity to generalize across a wide range of inputs, reinforcing its potential for applications in video generation tasks requiring high fidelity and adaptability.

\clearpage
\newpage
\bibliography{reference}
\bibliographystyle{apalike}

\end{document}